\documentclass[lettersize,journal]{IEEEtran}
\usepackage{amsmath,amsfonts}
\usepackage{algorithmic}
\usepackage{algorithm}
\usepackage{array}
\usepackage[caption=false,font=normalsize,labelfont=sf,textfont=sf]{subfig}
\usepackage{textcomp}
\usepackage{stfloats}
\usepackage{url}
\usepackage{verbatim}
\usepackage{graphicx}
\usepackage[table]{xcolor}
\usepackage{cite}
\usepackage{booktabs}
\usepackage{caption}
\usepackage{subcaption}  

\usepackage{multirow}
\usepackage{arydshln}
\usepackage{pifont}
\usepackage{makecell}
\begin{document}
\newcommand{\papertitle}{Understanding Multimodal Complementarity for Single-Frame Action Anticipation}

\title{\papertitle}

\author{Manuel Benavent-Lledo, Konstantinos Bacharidis, Konstantinos Papoutsakis,\\Antonis Argyros, and Jose Garcia-Rodriguez
\thanks{M.~Benavent-Lledo and J.~Garcia-Rodriguez are with the Department of Computer Technology, University of Alicante, Carr. de San Vicente del Raspeig s/n, E-03690 San Vicent2 del Raspeig, Alicante, Spain\\
E-mail: \{mbenavent,jgarcia\}@dtic.ua.es

K.~Bacharidis, K.~Papoutsakis and A.~Argyros are with the Institute of Computer Science, FORTH, Heraklion GR-700 13, Greece.\\
E-mail: \{kbach,papoutsa,argyros\}@ics.forth.gr

K.~Bacharidis and A.~Argyros are with the Computer Science Department, University of Crete, Heraklion, GR-700 13, Greece.

K.~Papoutsakis is with the Department of Management, Science and Technology, Hellenic Mediterranean University, Heraklion, GR-714 10, Greece.
}}% <-this % stops a space

        % <-this % stops a space
%\thanks{This paper was produced by the IEEE Publication Technology Group. They are in Piscataway, NJ.}% <-this % stops a space
%\thanks{Manuscript received April 19, 2021; revised August 16, 2021.}}

% The paper headers
%\markboth{Journal of \LaTeX\ Class Files,~Vol.~14, No.~8, August~2021}%
%{Shell \MakeLowercase{\textit{et al.}}: A Sample Article Using IEEEtran.cls for IEEE Journals}

%\IEEEpubid{0000--0000/00\$00.00~\copyright~2021 IEEE}
% Remember, if you use this you must call \IEEEpubidadjcol in the second
% column for its text to clear the IEEEpubid mark.

\maketitle
\begin{abstract}
Human action anticipation is commonly treated as a video understanding problem, implicitly assuming that dense temporal information is required to reason about future actions. In this work, we challenge this assumption by investigating what can be achieved when action anticipation is constrained to a single visual observation. We ask a fundamental question: how much information about the future is already encoded in a single frame, and how can it be effectively exploited?
Building on our prior work on Action Anticipation at a Glimpse (AAG), we conduct a systematic investigation of single-frame action anticipation enriched with complementary sources of information. We analyze the contribution of RGB appearance, depth-based geometric cues, and semantic representations of past actions, and investigate how different multimodal fusion strategies, keyframe selection policies and past-action history sources influence anticipation performance. 
Guided by these findings, we consolidate the most effective design choices into AAG+, a refined single-frame anticipation framework. Despite operating on a single frame, AAG+ consistently improves upon the original AAG and achieves performance comparable to, or exceeding, that of state-of-the-art video-based methods on challenging anticipation benchmarks including IKEA-ASM, Meccano and Assembly101. Our results offer new insights into the limits and potential of single-frame action anticipation, and clarify when dense temporal modeling is necessary and when a carefully selected glimpse is sufficient.
\end{abstract}


\section{Introduction}

\IEEEPARstart{H}{uman} action anticipation aims to predict an upcoming action before it occurs. This capability is fundamental to proactive perception systems, enabling early, safe and effective planning in a wide range of real world scenarios such as autonomous driving~\cite{rasouli2020pedestrianactionanticipationusing,liu2020spatiotemporal,girase2021loki}, procedural industrial tasks~\cite{Sener_2022_CVPR,Ben-Shabat_2021_WACV} or human-robot collaboration~\cite{jang2020etri,hwang2021eldersim,dai2022toyota}. Rather than acting reactively, anticipation allows systems to prepare for what is about to happen and adjust their behavior accordingly.
In structured procedural tasks such as industrial assembly and maintenance, action anticipation is particularly valuable. Anticipating the next step enables downstream decision making, reduces interruptions, and enables early detection of unsafe or erroneous transitions, supporting both productivity and safety.

\begin{figure}[t]
    \centering
    \includegraphics[width=\linewidth]{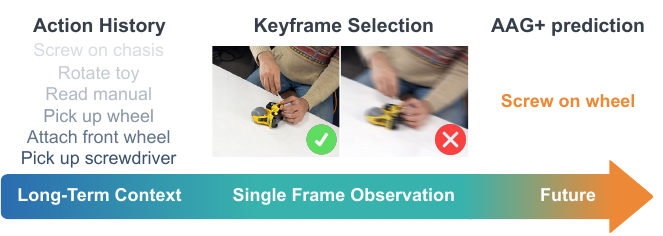}
    \caption{\textbf{Do we need full video for action anticipation, or does a single frame suffice?} By fusing a semantic action history with an informative frame, our model matches in performance that of dense video-based approaches. In this work, we also study the roles of various modalities, fusion strategies, and choices of action history source and keyframe selection methods in overall performance.}
    %\caption{How much information is necessary to anticipate the next action in a complex assembly task? Is a single-frame observation sufficient, or do geometric cues and long-term context improve prediction? We show that fusing a single RGB frame with complementary visual modalities and semantic history of past actions enables accurate action anticipation without relying on dense video processing. We explore (a) the contribution of each modality, (b) how they should be fused, and (c) whether selecting the frame using a key-frame selection policy leads to better performance.}
    \label{fig:concept-fig}
\end{figure}

Most contemporary anticipation methods approach the problem as dense video understanding~\cite{lai2024human,AVT,TempAgg,RULSTM}, using short- and long-range temporal aggregation to infer intent from motion and evolving context. More recent approaches further incorporate multimodal cues (e.g., audio, pose, or object interactions)~\cite{Manousaki_2023_ICCV,zhao2023antgpt,zatsarynna2023action,plausiVL2024,cao2025vision,guo2025towards,zhang2024object}, increasing the amount of information processed per time step. While effective, these strategies often require processing vast amounts of temporally correlated data and learning temporal dynamics over extended horizons, increasing computational cost and modeling complexity.
However, a key observation is that temporal redundancy is often high~\cite{kodathala2025temporal,doorenbos2025video,yang2025video,zhou2025glimpse,lei2022revealing}. This property is particularly pronounced in procedural and instructional settings, where anticipation-relevant evidence often appears in brief moments that reveal object states, grasp configurations, and intermediate completion signals.  Such observations can be interpreted as priors that reveal constrain or impose preconditions to plausible action transitions. This insight motivates us to seek for an alternative paradigm, that of {\em single-frame anticipation} (Fig.~\ref{fig:concept-fig}), in which anticipation is performed from a single observation augmented with contextual cues rather than from dense temporal encoding.

Single-frame anticipation is a challenging problem because without additional priors, a single image may reveal only partial task progress and intent.  Consequently, the key question is not whether a single frame alone can match video-based modeling, but how to combine appearance with contextual priors to recover much of the benefit of temporal reasoning that enables state-of-the-art performance, while reducing its modeling complexity. In our previous work~\cite{aag}, we introduced Action Anticipation at a Glimpse (AAG), a framework demonstrating that a single RGB frame, when augmented with additional modalities, can compete with video-based methods for action anticipation. Specifically, we showed that combining RGB appearance, geometric cues from depth, and semantic summaries of previously executed actions, enables accurate anticipation without relying on dense video processing. While these results were promising, they left several important questions unanswered regarding \emph{how} and \emph{why} such multimodal single-frame representations succeed.

In this work, we set out to extend this investigation and address these questions by systematically studying multimodal complementarity in the single-frame regime under different multimodal fusion strategies. We analyze when RGB appearance is sufficient, when depth-derived geometric information provides consistent gains, and when semantic priors from action history dominate the prediction. Beyond the presence of each modality, we show that the fusion mechanism critically influences whether modalities reinforce or interfere with each other, and therefore we evaluate multiple fusion strategies.

We further consider two deployment-relevant factors. First, not all frames are equally informative: blur, occlusion, and viewpoint can severely affect performance, as observed in similar video understanding tasks~\cite{sun2025frames,liang2024keyvideollm}. Accordingly, we study keyframe selection policies and their impact across different viewpoints. Second, action history priors are inherently noisy and imperfect, as they are derived from auxiliary predictive models. To address these limitations, we introduce a refinement mechanism that enhances robustness in practical settings through a stochastic corruption operator.

In this context, an additional question is how to obtain the semantic action history used as contextual prior. Recent progress in large language models (LLMs) and vision-language models (VLMs) suggests the possibility of inferring procedural structure or past actions directly from a single image~\cite{ahn2025happens,zhou2025glimpse}. Motivated by the observation that humans can often infer plausible prior steps from a snapshot in structured environments, we compare histories inferred by modern VLMs~\cite{team2023gemini,singh2025openai} with those produced by a single-frame action recognition model. Furthermore, we evaluate VLM-based histories under varying levels of prompt specificity and contextual information, in line with recent findings on the benefits of fine-grained prompting~\cite{wang2025action,yuan2025video,zou2025unlocking}.

%\textcolor{cyan}{In this context, another interesting topic to research is to exploit the recent progress in LLMs and VLMs to encode action history or extract goal-directed structure for anticipation. Motivated by the observation that humans can often infer plausible prior steps from a single image in structured environments, we empirically evaluate whether modern commercial VLMs can reliably reconstruct fine-grained procedural histories from a single frame. Our extended study—covering newer VLM versions and a wide range of prompt refinement strategies—shows that, while VLM outputs are typically fluent and plausible, they frequently hallucinate, omit, or reorder critical actions in procedurally complex scenarios, yielding histories that are inconsistent with the true task state. Consistent with this finding, histories populated by an auxiliary action recognizer (AR) are temporally grounded and substantially more reliable, translating into improved anticipation performance; accordingly, AAG+ adopts AR-based history population as its default configuration. This design choice is also reflected in our experimental evaluation, where AAG+ achieves consistent gains over AAG under the single-frame regime across IKEA-ASM and Meccano, and improves the primary metric on Assembly101, while avoiding the instability introduced by VLM-inferred histories.}

Based on these findings, we propose AAG+, an enhanced multimodal single-frame framework that consistently improves upon AAG through more robust modality selection and more effective fusion strategies, building on insights from multimodal transformer architectures~\cite{10123038}, while preserving its efficiency advantages. Additionally, it achieves competitive performance with state-of-the-art video-based methods. Our contributions are summarized as follows:

\begin{itemize}
    \item We conduct a systematic analysis of modality contributions in single-frame action anticipation, quantifying the impact of RGB appearance, depth-based geometric cues, and semantic past-action context.
    \item Building on these insights, we study multimodal fusion and frame-selection strategies and introduce AAG+, which combines the most effective modalities and design choices, yielding consistent improvements over AAG and competitive performance with state-of-the-art video-based methods.
    \item We further study the role of past action context in single-frame anticipation, comparing histories derived from an action recognition model with those inferred by VLMs under different prompts and levels of contextual information, analyzing their temporal consistency and reliability.

    %\item We investigate a range of \textbf{multi-modal fusion strategies}, analyzing how different fusion mechanisms affect anticipation accuracy and robustness.
    %\item We study the role of \textbf{key-frame selection policies}, examining whether and how frame selection influences single-frame anticipation performance.
    %\item 
    % We consolidate our findings into an enhanced version of the AAG framework, combining the most effective modalities, fusion strategies, and key-frame selection policies, achieving improved performance over the original AAG and comparable or superior results to state-of-the-art video-based methods.
    %We introduce AAG+, consolidating the strongest modalities, fusion mechanisms, and frame-selection policies, and demonstrate improved performance over AAG with competitive results against state-of-the-art video-based methods.
    %\item We also provide an extended evaluation of VLM-based action-history inference (including prompt refinement), and show that VLM predictions on the past and the next anticipated action are often inconsistent with the procedural state, under-performing temporally grounded action histories populated by action recognition toward next action anticipation. +++
    \end{itemize}

All findings are validated through extensive experiments on diverse procedural activity datasets (IKEA-ASM~\cite{Ben-Shabat_2021_WACV}, Meccano~\cite{RAGUSA2023103764}, Assembly101~\cite{Sener_2022_CVPR}), providing practical insights into when multimodal single-frame anticipation can serve as an effective alternative to dense video-based approaches. The implementation of AAG+ will be made publicly available\footnote{\url{https://github.com/ManuBenavent/AAG-Plus}}. % on Github - 

%The remainder of the paper is organized as follows. Section~\ref{sec:related} reviews related work on action anticipation and.... Section~\ref{sec:method} details the proposed AAG+ framework. 

The remainder of the paper is organized as follows. Section~\ref{sec:related} reviews related work. Section~\ref{sec:method} details the proposed framework design. Section~\ref{sec:setup} describes the datasets and evaluation protocol. Section~\ref{sec:results} reports the main results and ablations on modalities, fusion, keyframe selection, and robustness to imperfect histories. Section~\ref{sec:conclusions} draws the main conclusions of this work.
\section{Related Work}
\label{sec:related}

\noindent We review existing approaches for multimodal video-based action anticipation, and subsequently concentrate on methods for single-frame action recognition and anticipation, due to the limited work in this area.
\vspace{0.1cm}

\noindent \textbf{Video-based Human Action Anticipation} has attracted significant attention in recent years~\cite{lai2024human}. Several approaches operate solely on RGB video inputs and achieve state-of-the-art results by leveraging effective temporal modeling strategies. Among them, TempAgg~\cite{TempAgg} proposes a versatile multi-scale temporal aggregation scheme aimed at tackling the complexities of long-range video understanding. Similarly, the Anticipative Video Transformer (AVT)~\cite{AVT} adopts an end-to-end attention-driven framework that simultaneously anticipates future actions and learns frame representations optimized for predicting forthcoming content. By explicitly maintaining the order of observed actions, AVT captures long-term temporal dependencies in a structured and principled way.

In contrast, multimodal methods leverage complementary information from multiple modalities to build more informative representations ahead of the classification stage. For example, in RULSTM~\cite{RULSTM}, authors propose a Rolling-Unrolling LSTM for action anticipation. A Rolling LSTM encodes observed frames to capture temporal structure and an Unrolling LSTM forecasts future actions and object interactions. Textual cues, typically expressed as action descriptions, have also been shown to enhance anticipation. The Visual-Linguistic Modeling of Action History framework (VLMAH)~\cite{Manousaki_2023_ICCV} integrates immediate visual features with a lightweight linguistic representation of past actions, enabling the modeling of both short-term and long-term context.

% AntGPT~\cite{zhao2023antgpt} uses large language models to extract goal-directed activities, also referred to as high-level actions, which improve the prediction of fine-grained actions. A similar strategy is used in~\cite{zatsarynna2023action}, where a joint loss enforces consistency between fine- and coarse-grained actions.
The task of action anticipation has recently seen a paradigm shift with the integration of large language models and vision-language models to provide semantic priors and high-level reasoning. AntGPT~\cite{zhao2023antgpt} leverages the reasoning capabilities of large language models to extract goal-directed activities, often termed high-level actions, from video sequences. By treating these high-level goals as anchors, the model improves the prediction of upcoming fine-grained actions through a more structured understanding of intent. A similar strategy is employed in~\cite{zatsarynna2023action}, where a joint loss function is introduced to enforce consistency between fine-grained future actions and coarse-grained goal predictions, ensuring that anticipated movements align with the overall objective of the actor. PlausiVL~\cite{plausiVL2024} extends the generative capabilities of video-language models by introducing a ``plausibility'' loss using counterfactual learning to distinguish between realistic and unrealistic action sequences, effectively acting as a semantic filter for future predictions. Moreover, the model proposed in~\cite{cao2025vision} explicitly infers behavioral intentions as textual features before feeding them into an LLM for final sequence forecasting. On the other hand, BiAnt~\cite{sato2025bidirectional} introduces bidirectional action sequence learning by training the model to predict both forward (past to future) and backward (future to past), enforcing a stricter form of temporal continuity and goal-consistency. Finally, ActionLLM~\cite{wang2025actionllm} further optimizes this paradigm by leveraging the role of VLMs as action anticipators. It treats video sequences as successive tokens, utilizing ``action-tuning'' to align the LLM’s internal knowledge with the specific constraints of the action anticipation domain.
\vspace{0.1cm}

\noindent \textbf{Single-Frame Human Action Recognition} remains a fundamentally difficult problem, primarily because the visual evidence is restricted to a single observation and thus lacks the temporal dynamics that typically disambiguate human activities. To compensate for this limitation, prior work has explored complementary strategies, including knowledge distillation, explicit contextual reasoning, and stronger feature learning mechanisms. Early efforts, such as \cite{gkioxari2015contextual}, employ region-based CNN formulations that couple person-centric regions with informative environmental cues and highlight that accurate recognition often depends on jointly modeling the actor and the surrounding scene. 
% is inherently challenging due to the absence of temporal cues. To address this, various methods leverage knowledge distillation, contextual reasoning, and enhanced feature extraction. The work in~\cite{gkioxari2015contextual} emphasizes the importance of jointly modeling actors and their environment using region-based CNNs, demonstrating how context significantly improves recognition performance.
More recent methods emphasize spatial reasoning and structured representation learning to better capture human-object interactions and scene semantics. Ashrafi et al.~\cite{ashrafi2023still} explicitly model relationships between body joints and manipulated objects, aiming to represent interaction patterns that are indicative of the underlying action. Liang et al.~\cite{liang2024patch} introduce a patch-based formulation that localizes actions in still images without requiring bounding-box supervision. In parallel, transformer-based architectures have been adopted to strengthen global reasoning over poses and contextual cues, leveraging vision transformers to extract spatial relations between human pose configurations and scene context~\cite{hosseyni2024human}. Model efficiency and generalization have also been improved in~\cite{saleknia2024multi}, by proposing progressive knowledge distillation to transfer discriminative capacity into lighter models. Finally, other approaches, including \cite{he2023context}, further enhance contextual reasoning in still images by combining pixel-level and region-level attention mechanisms.
\noindent \textbf{Single-Frame Human Action Anticipation} has been relatively underexplored despite its computational advantages~\cite{10.1007/978-3-319-10602-1_28}, including avoiding temporal aggregation and enabling stronger spatial feature learning when studied alongside video supervision \cite{lei2022revealing}. Owing to the limited context available in a single image, existing solutions are largely multimodal. Lan et al.~\cite{lan2014hierarchical} introduce hierarchical ``movemes'', a multi-level representation that captures human movement at different granularities to anticipate future actions from a single image or short clip. Vondrick et al.~\cite{vondrick2016anticipating} propose a self-supervised model that predicts high-level future visual features from a single image and produces diverse outcomes to account for uncertainty.
Recent vision-language approaches emphasize common-sense-driven contextualization rather than explicit action prediction. VisualCOMET~\cite{park2020visualcomet} infers dynamic context by reasoning about prior events, intentions, and plausible future outcomes. Similarly, ActionCOMET~\cite{sampat2024actioncomet} derives action-related common-sense knowledge (e.g., preconditions, outcomes, intentions, and sequences) using language models and annotated cooking videos. Although neither VisualCOMET nor ActionCOMET performs direct action anticipation, both provide semantic signals that can support downstream reasoning tasks.
Finally, our previous work on action anticipation introducing AAG~\cite{aag} explicitly addresses the challenge of predicting upcoming actions from a single moment. Unlike video-based models that require temporal aggregation, AAG leverages a single RGB frame enriched with depth cues and textual summaries to provide long-term context. Results demonstrate that semantic priors can compensate for the lack of temporal visual information.

\section{Multimodal Refinement for Action Anticipation at a Glimpse}
\label{sec:method}

\begin{figure*}[t]
    \centering
    \includegraphics[width=\linewidth]{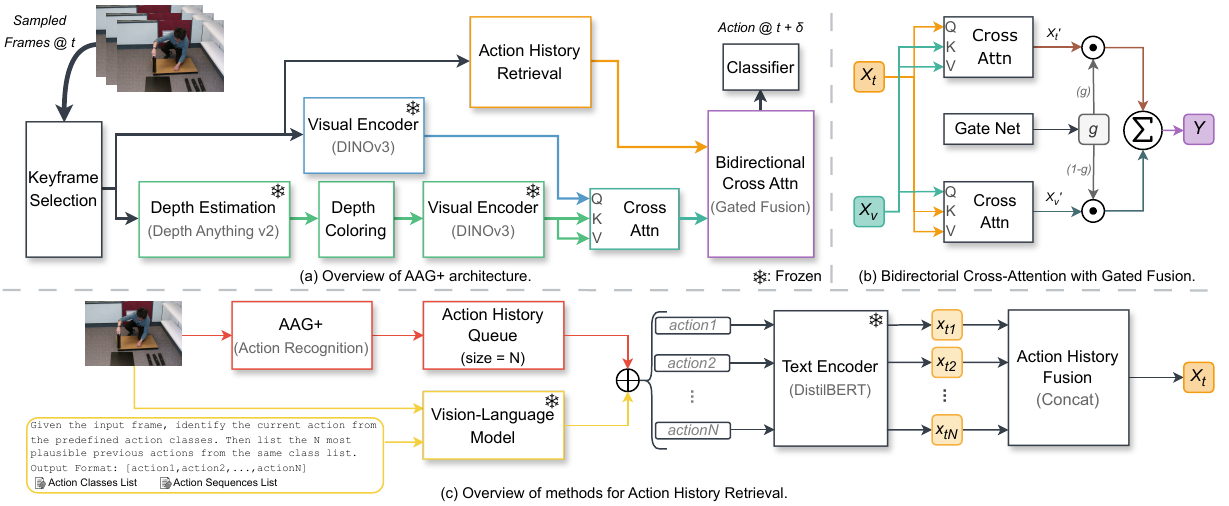}
    \caption{\textbf{Illustration of AAG+. (a)~Overview of AAG+ architecture.} Following keyframe selection, three modalities are processed. Depth (green) is estimated and colorized prior to feature extraction and fusion with RGB features (blue). Visual fusion is performed via cross-attention, where RGB acts as the primary modality. Visual features are integrated with textual context (orange) extracted from the action history retrieval module through bidirectional cross-attention (purple). 
   \textbf{(b)~Bidirectional Cross-Attention with Gated Fusion.} Input features undergo independent cross-attention and are then combined through a gated fusion, where a learned gating vector $g$ adaptively weights the contribution of each modality prior to summation.
   \textbf{(c)~Overview of methods for Action History Retrieval.} 
   We propose two methods for history retrieval: (yellow) prompting a VLM to predict the past actions sequence, or (red) repurposing the AAG+ architecture to perform action recognition on the current frame and storing the results in a queue of size $N$. In both cases, class names are mapped to a latent space via a text encoder to produce individual embeddings $x_{tn}$, which are subsequently fused into the joint representation $X_t$.}
    \label{fig:method}
\end{figure*}

% \noident: common in IEEE template on the start of every section
\noindent We revisit the core design of Action Anticipation at a Glimpse and present an enhanced version, AAG+, which addresses the limitations identified in the original framework. AAG+ refines modality processing through improved feature extractors, introduces a keyframe selection mechanism, and incorporates a stochastic corruption operator to handle imperfect action histories. We further provide a focused analysis of multimodality and fusion strategies, with all extensions guided by a systematic evaluation of individual component and modality contributions.
%(a) \textit{refinements in modality processing}, (b) \textit{advanced multi-modal fusion strategies}, and (c) \textit{a frame selection process}, with all novel additions/improvements being informed by a deeper analysis of each component contribution. %Rather than listing all possible variants, we focus on the best-performing configurations identified in our extended study, with full ablation results provided in Section~\ref{sec:experiments}.

To structure the presentation, we organize it around the three main modules of AAG+: the visual encoder module, the action history retrieval module, and the multimodal fusion and classification module. In the following sections, we revisit each module, describe the original design, and highlight the specific refinements introduced in AAG+ to overcome the original limitations. An overview of the AAG+ framework is presented in Figure~\ref{fig:method}.

%%%------------ SubSection----------------------
\subsection{Visual Encoder}

\noindent In the AAG framework, visual reasoning is constrained to a single observation, under the assumption that a single frame can encode sufficient information about the scene state when enriched with complementary modalities. Specifically, the encoder uses an RGB frame, $f_t$, and an aligned depth representation, $d_t$, from the last observed moment $t$ to predict actions $\delta$ seconds into the future. RGB appearance features are extracted using a self-supervised vision transformer, originally DINOv2~\cite{oquab2023dinov2}, yielding $X_{\text{RGB}} \in \mathbb{R}^{D_{vis}}$, where $D_{vis}$ denotes the feature dimensionality of the encoder. RGB features capture object appearance, layout, and actor-object interactions, but lack explicit geometric structure.

To complement appearance cues, AAG incorporates depth information, which provides explicit spatial relationships between scene elements. Since depth maps are single-channel and often noisy, depth is first estimated using a monocular depth model (Depth Anything v2~\cite{NEURIPS2024_26cfdcd8}) and converted into a pseudo-RGB representation via color mapping similar to~\cite{RAGUSA2023103764}. This enables the use of the same feature extractor as for RGB, producing a depth embedding $ X_{\text{D}} \in \mathbb{R}^{D_{vis}}$.

To integrate the two modalities, a cross-attention mechanism is employed in which RGB features act as queries and depth features as keys and values:
\begin{equation}
    Q = W_Q X_\text{RGB}, \qquad
    K = W_K X_\text{D}, \qquad
    V = W_V X_\text{D},
    \label{eq:qkv_ca}
\end{equation}
followed by the scaled dot-product attention operator:
\begin{equation}
    X_V = \text{softmax}\left(\frac{QK^\top}{\sqrt{D}}\right)V,
    \label{eq:attention}
\end{equation}
where $W_Q, W_K, W_V \in \mathbb{R}^{D \times D}$ are learnable projections and $D$ denotes the shared embedding dimension. This formulation allows depth cues to selectively refine the RGB representation, emphasizing spatially informative regions while preserving appearance-driven semantics.

In AAG+, we maintain this cross-attention strategy to fuse visual features. This decision is informed by previous results demonstrating that this approach outperforms alternatives such as self-attention. It is also supported  by the multimodal analysis conducted in Section~\ref{sec:results}, which reveals that limitations in visual reasoning reside in the input feature quality rather than in the fusion strategy itself.

Particularly, while the visual encoder proved effective in controlled exocentric settings (e.g. IKEA-ASM), we observe clear limitations in more challenging scenarios, where increased camera motion induces blur and frequent occlusions reduce object visibility, a common issue in egocentric or close-up view perspectives (e.g. Meccano, Assembly101). Under such conditions, the quality of the single observed frame becomes a dominant factor: noisy depth estimates and visually degraded RGB frames can overwhelm the cross-attention mechanism, limiting its ability to provide meaningful refinement.

\vspace{0.1cm}
\noindent\textbf{Keyframe Selection.}
Motivated by these observations, we introduce refinements to the visual encoder that explicitly account for frame quality. Rather than strictly sampling the last frame, we adopt a keyframe selection strategy that prioritizes frames with low motion blur and high visual clarity. By ensuring that both RGB and depth features are extracted from visually informative frames, the cross-attention mechanism operates on more reliable inputs, leading to improved robustness across datasets. 

The keyframe selection module explicitly accounts for frame quality within a short temporal window into the past, set to five frames in our experiments. We evaluate three distinct keyframe selection strategies:

\begin{itemize}
\item \textbf{Blurriness-based Selection (Laplacian Variance):} We estimate the focus of each frame using the variance of the Laplacian. At each time step $t$, we examine the previous five frames, starting from the most recent frame $f_t$ and moving backward. We select the first frame in this window whose Laplacian variance exceeds a predefined sharpness threshold. Should none of the frames meet the threshold, we default to $f_t$. Empirically, Laplacian variance over $300$ indicates very sharp frames, $100 - 300$ indicates sharp frames, $50 - 100$ moderate blur, and under $50$ pronounced motion blur. Thresholds are chosen per dataset based on viewpoint characteristics.

\item \textbf{Distance-based Selection ($L_2$ and Cosine):} For each time step $t$, we extract RGB embeddings for all frames within the previous short-term window. We then compute the centroid of these embeddings and select the frame closest to this centroid as the most representative. This approach prioritizes frames that are semantically stable and representative of the recent temporal context.

\end{itemize}

AAG+ adopts blurriness-based keyframe selection as it outperforms distance-based strategies (see Section~\ref{sec:results}) while significantly reducing computational overhead, as no feature extraction is required for the candidate frames.

\vspace{0.1cm}
\noindent\textbf{Feature Extraction.}
In addition, we revisit the choice of visual backbone, which plays a critical role in the single-frame anticipation setting. When temporal redundancy is absent, the quality and robustness of the extracted visual representation directly constrain the model’s ability to reason about scene state and future actions. We therefore replace DINOv2~\cite{oquab2023dinov2} with DINOv3~\cite{simeoni2025dinov3}, which offers stronger representations and improved robustness to noise, motion blur, and appearance variation, which are factors that are particularly pronounced in unconstrained assembly scenarios.

We further explored more specialized depth-aware architectures, including Depth Anything 3~\cite{lin2025depth}. However, despite their suitability for dense geometric reasoning, these alternatives did not yield consistent improvements for single-frame anticipation compared to the one used in AAG, Depth Anything~v2~\cite{NEURIPS2024_26cfdcd8}. As such, they are not adopted in the final model. Ablation results on the visual feature encoders are provided in Appendix~A.

%%%------------ SubSection----------------------
\subsection{Action History Retrieval}
\noindent Action anticipation is inherently ambiguous, particularly when relying on a single visual observation. In the absence of explicit temporal modeling, information about previously executed actions provides crucial context for disambiguating plausible futures. AAG addressed this challenge by representing past actions in textual form, motivated by prior work showing that long-term temporal dependencies can be effectively encoded using language-based representations~\cite{Manousaki_2023_ICCV,benaventlledo2024enhancing}.

Building on this insight, within the AAG framework we explored using a single-frame action recognizer to obtain past actions, encoded either as free-form textual descriptions or as individual per-action embeddings. Experiments showed that representing each past action independently and aggregating the embeddings consistently outperformed encoding the entire history as a single summary. Motivated by these results and the ablations in AAG, we focus on independent action encodings aggregated via concatenation. This strategy is both more effective than alternative aggregation methods (e.g., self-attention), and computationally efficient: since action class names are known in advance, embeddings can be precomputed, avoiding the need to run a text encoder at inference time.

Formally, let $x_{tn} \in \mathbb{R}^{D_\text{txt}}$ denote the embedding of the $n$-th past action at time step $t$, obtained from the text encoder. The joint action history representation at time $t$ is then computed as the concatenation of all $N$ past action embeddings:
\begin{equation}
    X_t = \text{Concat}(x_{t1}, x_{t2}, \dots, x_{tN}) \in \mathbb{R}^{N \cdot D_\text{txt}}.
\end{equation}
The number of past actions $N$ is determined empirically through ablation studies per dataset. The resulting representation $X_t$ is subsequently projected into the joint embedding space, $D$, to be combined with visual features.

In AAG+, we retain this strategy and use DistilBERT~\cite{sanh2019distilbert} as the text encoder and explore two key extensions: (1) the source of action history and (2) robustness to imperfect histories. While AAG considered vision-language models producing textual descriptions, a direct comparison with independent action embeddings has not been studied. We evaluate recent VLMs~\cite{singh2025openai,team2023gemini} and fine-grained prompting strategies~\cite{wang2025action,yuan2025video,zou2025unlocking} for this purpose. Additionally, since action histories are prone to noise due to reliance on auxiliary models (single-frame recognizers or VLMs), we introduce mechanisms to improve robustness under such conditions.

\vspace{0.1cm}
\noindent\textbf{Action History Source.}  
We evaluate two primary sources for constructing the action history:

\begin{itemize}
    \item \textbf{VLM-based inference:} Inspired by the observation that humans can often infer prior steps from a single moment, particularly in structured environments, VLMs are used to reconstruct plausible past actions from a single frame. We evaluate state-of-the-art vision-language models (GPT~5.1~\cite{singh2025openai} and Gemini~3~\cite{team2023gemini}) under varying prompts, which differ in the level of detail in the instructions and the amount of contextual information provided, including class names, action sequences, and frequency of prior steps, among others. The number of past actions $N$ is explicitly specified in the prompt, and model outputs are parsed into individual action class names. An example is shown in Figure~\ref{fig:method}.c, while full prompt specifications and qualitative results are provided in Appendix~E.
    \item \textbf{Single-frame action recognizer}: Past actions are obtained from predictions of an action recognition model applied to the current frame, maintaining a queue of previously completed actions of size $N$ under realistic inference conditions. This allows reuse of features already computed for action anticipation, providing an efficient, temporally grounded, and consistent action history representation.
\end{itemize}

\noindent\textbf{Robustness to Imperfect Histories.}
No action history constructing method is immune to errors. To mitigate over-reliance on perfectly ordered ground-truth sequences observed during training, AGG+ introduces robustness mechanisms that encourage the model to handle imperfect histories. Specifically, we define a stochastic corruption operator $\mathcal{C}(\cdot)$ that transforms the joint history embedding $X_T$ into a perturbed version $\tilde{X}_T$:  

\begin{equation}
    \tilde{X}_T = \mathcal{C}(X_T),
\end{equation}

where $\tilde{X}_T$ is used during training in place of $X_T$. Two instantiations of $\mathcal{C}$ are considered:  

\begin{itemize}
    \item \textbf{Random embedding noise}: To increase robustness to small inaccuracies in the action history, the action history embedding, $X_T$, is stochastically perturbed during training using a combination of dropout and Gaussian noise. Each element of $X_T$ is independently zeroed with probability $p$ and rescaled by $1/(1-p)$:
    \begin{equation}
    \tilde{X}_T \leftarrow X_T \odot m / (1-p), \quad m_i \sim \text{Bernoulli}(1-p),
    \end{equation}
    where $\odot$ denotes element-wise multiplication.  
    
    Subsequently, Gaussian noise with standard deviation $\sigma$ is added to the embeddings:
    \begin{equation}
    \tilde{X}_T \leftarrow \tilde{X}_T + \epsilon, \quad \epsilon \sim \mathcal{N}(0, \sigma^2 \mathbf{1}),
    \end{equation}
    where $\mathbf{1}$ is a vector of ones matching the dimensionality of $X_T$. During the experiments, $\sigma$ was set to 0.03.
    
    \item \textbf{Action swapping}: To simulate realistic errors in the action history rather than applying random noise to the whole embedding, each action in the sequence is independently replaced, with probability $p$, by a randomly selected alternative action from the dataset. Let $x_{tn}$ denote the embedding of the $n$-th action, then, for each action:
    \begin{equation}
        x_{tn} \leftarrow 
        \begin{cases}
            x_{tn}', & \text{with probability } p, \\
            x_{tn}, & \text{with probability } 1-p,
        \end{cases}
    \end{equation}
    where $x_{tn}'$ is the embedding of a randomly selected alternative action. This mechanism introduces both incorrect predictions and potential ordering inconsistencies while preserving the per-action embedding structure. Notably, the probability $p$ is closely linked to the complexity of the dataset, the number of previous actions considered, $N$, and the accuracy of the action history retrieval method. Accordingly, we perform ablations on $p$ for each dataset to identify the most effective settings.
\end{itemize}

%%%------------ SubSection----------------------
\subsection{Multimodal Fusion \& Classification}
\noindent Visual and textual cues are fused to form a global representation, $Y$, which is used for classification. In AAG, this fusion is performed using a self-attention mechanism, which was shown to outperform simpler strategies based on concatenation or element-wise arithmetic operations. However, the multimodal analysis presented in Section~\ref{sec:results} reveals limitations in the ability of this mechanism to effectively integrate visual and textual embeddings.

For AAG+, we conducted an extensive study of attention-based fusion mechanisms to combine visual and textual feature spaces. Specifically, we evaluated cross-attention with residual and gated variants, bidirectional cross-attention with shared or independent weights, and alternative strategies for combining the resulting representations (e.g., concatenation, element-wise operations, self-attention, or gated fusion). Rather than listing all variants here, we focus on the top-performing configuration: bidirectional cross-attention with gated fusion, which integrates modality-specific representations through a learned gating mechanism (Fig.~\ref{fig:method}.b). Full descriptions and detailed comparisons of all variants are provided in Appendix~A.

In this scheme, cross-modal interactions are extended beyond a single direction, allowing each modality to attend to the other. Formally, given visual and textual feature sequences $X_v$ and $X_t$, we compute cross-attended representations in both directions by applying the cross-attention mechanism defined in Equations~\ref{eq:qkv_ca}-\ref{eq:attention}:

\begin{equation}
\begin{aligned}
X_v' &= \text{CA}(Q = X_v, K = X_t, V = X_t),\\
X_t' &= \text{CA}(Q = X_t, K = X_v, V = X_v),
\end{aligned}
\end{equation}
where $\text{CA}(\cdot)$ denotes the cross-attention operation described previously. The resulting representations $X_v'$ and $X_t'$ are then combined via a gated fusion:

\begin{equation}
Y = g \odot X_v' + (1 - g) \odot X_t',
\end{equation}
where $g \in [0,1]$ is a learned gating vector that adaptively weights the contribution of each modality. This mechanism enables the model to integrate complementary information from both streams while modulating their influence according to reliability or relevance. The resulting representation $Y \in \mathbb{R}^D$ serves as the global feature for AAG+ and is fed to an MLP for classification using cross-entropy loss for supervision.

\section{Experimental Setup}
\label{sec:setup}

\subsection{Datasets}
\noindent We evaluate our approach on three publicly available datasets designed for action recognition and anticipation in industrial scenarios. They encompass diverse environments, action granularity levels, and procedural variability.
 
\textbf{IKEA-ASM}~\cite{Ben-Shabat_2021_WACV} contains third-person videos of furniture assembly featuring 33 atomic actions and 4 high-level activities. Following \cite{aag} we use the top RGB view and use the default environment-based train/test split.

\textbf{Meccano}~\cite{RAGUSA2023103764} is an egocentric dataset comprising videos of participants assembling a toy motorbike, recorded with head-mounted sensors. It includes 61 action classes derived from 20 object and 12 verb categories. Following prior work~\cite{aag, Manousaki_2023_ICCV, RULSTM}, we use the standard 11/9 train/test split.

\textbf{Assembly101}~\cite{Sener_2022_CVPR} contains videos of participants assembling and disassembling 101 toy vehicles without fixed instructions, resulting in diverse execution patterns. It includes over 1M action segments spanning 1380 fine-grained action classes, derived from 90 objects and 24 verbs, captured across 12 camera views. We use the exocentric (third-person) camera \emph{v4}, which provides the best RGB performance~\cite{Sener_2022_CVPR}, and follow the train/validation splits from~\cite{Manousaki_2023_ICCV, aag}, as the official test set is not publicly available.

\subsection{Methods \& Implementation Details}
\noindent \textbf{Evaluation:} We evaluate our method by predicting actions $\delta = 1s$ into the future and report performance using two standard metrics in action anticipation: top-$k$ accuracy and class-mean Top-5 recall~\cite{Manousaki_2023_ICCV,RAGUSA2023103764,Sener_2022_CVPR,damen2022rescaling}. Top-$k$ accuracy captures uncertainty in future predictions, while class-mean Top-5 recall addresses long-tail class distributions. Following prior work, we adopt top-$k$ accuracy as the primary metric for IKEA-ASM and Meccano~\cite{Manousaki_2023_ICCV,RAGUSA2023103764}, and class-mean Top-5 recall for Assembly101~\cite{Sener_2022_CVPR}. In the reported tables, bold font indicates the best performance for each dataset’s primary metric. To evaluate the role of action history, we report results using both ground-truth and predicted histories. 

Experiments involving action recognition report top-1 accuracy following established benchmarks~\cite{Ben-Shabat_2021_WACV,RAGUSA2023103764,Sener_2022_CVPR}.
\vspace{0.1cm}

\begin{table*}[t]
    \centering
    %\small
    \setlength{\tabcolsep}{3pt} % Reduce column spacing
    \caption{Comparison with state-of-the-art video-based action anticipation methods.\\ {\footnotesize Action History (AH) is predicted by an action recognizer (single-frame for AAG \& AAG+, video-based for VLMAH). Underline: best single-frame performance.}    }
    \label{tab:performance}
    \begin{tabular}{llrrr rr rr}
        \toprule
        \textbf{Method} & \textbf{Modalities} & \multirow{2}{*}{\makecell{\textbf{\# Input}\\\textbf{Frames}}} & \multicolumn{2}{c}{\textbf{IKEA-ASM}} & \multicolumn{2}{c}{\textbf{Meccano}} & \multicolumn{2}{c}{\textbf{Assembly101}} \\
        \cmidrule(lr){4-5} \cmidrule(lr){6-7} \cmidrule(lr){8-9}
        & & & Top-1/5 Acc & Recall@5 & Top-1/5 Acc & Recall@5 & Top-1/5 Acc & Recall@5 \\
        \midrule
        RULSTM~\cite{RULSTM} & RGB & 14 & 26.37 / 70.05 & 16.00 & 24.08 / 58.23 & 22.38 & 5.95 /21.28 & 1.00 \\
        TempAgg~\cite{TempAgg} & RGB & 37 & 26.90 / 70.14 & 16.42 & 19.69 / 25.37 & 17.47 & 11.43 / 34.16 & 9.07 \\
        AVT~\cite{AVT} & RGB & 10 & 27.10 / 69.70 & 45.78 & 27.43 / 53.38 & 32.05 & 7.25 / 23.80 & 17.03 \\
        VLMAH~\cite{Manousaki_2023_ICCV} & RGB, AH & 8 & \textbf{52.31 / 85.44} & 47.45 & \textbf{29.11 / 57.05} & 57.05 & 9.17 / 27.63 & \textbf{25.13} \\\noalign{\vskip 0.5ex}\hdashline\noalign{\vskip 0.5ex}
        AAG~\cite{aag} & RGB-D, AH & 1 & 44.66 / 82.87 & 46.59 & 26.43 / 54.95 & 12.27 & 13.46 / 32.87 & 9.80  \\
        \textbf{AAG+} & RGB-D, AH & 1 & \underline{51.26 / 88.88} & 54.78 & \underline{27.24 / 60.41} & 16.12 &   13.14 / 32.28 & \underline{13.02} \\
        \bottomrule
    \end{tabular}
    
\end{table*}

\noindent \textbf{Implementation Details:} We use DINOv3~\cite{simeoni2025dinov3} with ViT-7B/16 backbone pretrained on LVD-1689M and half-precision (FP16), and DistilBERT~\cite{sanh2019distilbert} as the visual and text encoders, respectively. Depth frames are estimated using Depth Anything v2 (Small)~\cite{NEURIPS2024_26cfdcd8}. Transformer encoders consist of 2 layers, 4 attention heads per layer, and an embedding dimension $D=768$.
As described in Section~\ref{sec:method}, AAG+ applies a cross-attention transformer for visual feature fusion, followed by a bidirectional cross-attention transformer with gated fusion for multimodal integration. Keyframe selection is based on Laplacian variance with a threshold of 100 on IKEA-ASM and Assembly101, and 50 for Meccano due to lower-quality frames (see Appendix~A for quantitative and qualitative comparisons of frame blurriness per dataset). Action history is retrieved using a single-frame action recognizer, as determined by the ablation studies in the next section. Even though full results on the accuracy of the single-frame action recognizer are provided in Appendix B, we report key finding here. Using an RGB frame and action history, the model achieves 72.38\% on IKEA-ASM, 33.19\% on Meccano, and 30.74\% on Assembly101. Adding depth yields: 72.95\%, 33.79\%, and 31.49\%, respectively.

For experiments involving action history (AH), ground-truth sequences are used during training, while predicted sequences are used at inference. Following \cite{aag}, the number of previous actions $N$ is set to 7 for IKEA-ASM and 3 for Meccano and Assembly101. The action swapping strategy is employed to simulate realistic errors, with swap probabilities $p$ of 0.1 and 0.4 on IKEA-ASM and Meccano, respectively. Random noise is used on Assembly101 with $p = 0.1$ and noise distribution set to $0.03$, as determined by the ablation study presented in the following section.
\vspace{0.1cm}

\noindent \textbf{Training Details:} AAG+ has been implemented using PyTorch, and experiments are conducted on Nvidia RTX 4090 GPUs. We employ AdamW optimizer with a weight decay of $0.01$ and a base learning rate of $5 \times 10^{-5}$. Models are trained for $100$ epochs, with an early stopping patience of $10$ epochs, and an improvement threshold of $0.001$. Batch size was set to 32.
\vspace{0.1cm}

\noindent \textbf{SOTA Methods Details:} We compare AAG+ with the original AAG framework, as well as four state-of-the-art video-based action anticipation methods, to assess the impact of single-frame versus video-based approaches. The methods include AVT~\cite{AVT}, RULSTM~\cite{RULSTM}, TempAgg~\cite{TempAgg}, and VLMAH~\cite{Manousaki_2023_ICCV}. Missing baseline results reported in the original papers are taken from \cite{aag}, which provides consistent train/test splits, camera views, and matched action history lengths.

\section{Experimental Results}
\label{sec:results}

%\subsection{Impact of Temporal Resolution on Action Anticipation: Single Frame vs. Video}
\subsection{Comparison to State-of-the-art Approaches}

\noindent Table~\ref{tab:performance} compares the proposed multimodal single-frame model AAG+ with its predecessor, AAG, and with video-based (multi-frame) state-of-the-art action anticipation methods.

Despite using only a single frame, AAG+ outperforms several video-based methods and achieves competitive performance relative to approaches that rely on attention-based temporal modeling, such as AVT, and multimodal methods such as VLMAH. In particular, on IKEA-ASM, AAG+ performs comparably to the top-performing method VLMAH and outperforms RGB-only video-based approaches including TempAgg, AVT, and RULSTM, highlighting that strong spatial features and contextual priors can partially substitute for explicit temporal aggregation in structured tasks.

On more complex procedural datasets such as Meccano and Assembly101, where action sequences are longer, more variable, and less predictable, AAG+ remains competitive with video-based methods. These results indicate that even under higher temporal uncertainty, single-frame models can exploit strong spatial cues and local context to anticipate plausible next actions. At the same time, a performance gap remains relative to video-based approaches that explicitly model long-term dependencies through computationally heavier temporal mechanisms, such as AVT, or multimodal video-based methods such as VLMAH, although AAG+ consistently outperforms methods relying on dense frame aggregation alone, including RULSTM and TempAgg. Our results highlight the potential of single-frame reasoning in structured and predictable environments, while also revealing its limitations, as temporal modeling continues to provide an advantage in more flexible or ambiguous tasks. We further compare computational complexity in Appendix C to illustrate the efficiency advantages of single-frame anticipation over video-based approaches.

%\textcolor{blue}{Despite using only a single frame, both AAG+ and AAG achieve competitive performance compared to video-based methods. On IKEA-ASM, AAG+ performs comparably to the top-performing video-based method (VLMAH), while outperforming TempAgg, AVT and RULSTM, highlighting that strong spatial features and contextual cues can partially substitute for temporal aggregation in structured tasks. On more complex procedural tasks, as the ones in Meccano and Assembly101, where action sequences are longer, variable, and less predictable, AAG+ remains competitive with video-based baselines. This suggests that even in settings with higher temporal uncertainty, single-frame models can exploit strong spatial cues and local context to anticipate plausible next actions, although a performance gap remains relative to video-based approaches that explicitly model long-term dependencies. Overall, these results highlight both the potential and limitations of single-frame reasoning: it can be sufficient in structured, predictable environments, but richer temporal modeling still provides an advantage in more flexible or ambiguous tasks.}

Compared to its predecessor, AAG+ consistently improves across all datasets. In addition to more robust visual features, refined keyframe selection mitigates AAG’s limitations with respect to viewpoint sensitivity. AAG+ also handles imperfect action histories more effectively through dedicated robustness mechanisms. Finally, AAG+ combines the available features more effectively, guided by the multimodal analysis presented in the next section.

%\textcolor{blue}{Compared to its predecessor AAG, AAG+ consistently improves across all datasets. The integration of more robust visual features, refined keyframe selection, and improved multimodal fusion leads to gains across structured and complex tasks. The results indicate that the combination of carefully selected keyframes, high-quality visual embeddings, and robustly fused action history helps the model resolve ambiguities even under challenging conditions. These improvements demonstrate that architectural enhancements can substantially boost single-frame action anticipation, narrowing the gap with video-based methods while retaining the computational efficiency inherent to the single-frame approach.}

\begin{table*}[t]
    \centering
    %\small
    \setlength{\tabcolsep}{3pt}
    \caption{Effect of each modality on AAG+ across datasets and action history source (GT vs. Predictions).\\
    {\footnotesize Underline: best per block. Bold: best under realistic settings. *: AH populated from AAG+'s action recognizer (RGB-D, AH).}}
    \label{tab:modalities}
    \begin{tabular}{ll rr rr rr}
        \toprule
        \textbf{AH Source} & \textbf{Modalities}
 & \multicolumn{2}{c}{\textbf{IKEA-ASM}} & \multicolumn{2}{c}{\textbf{Meccano}} & \multicolumn{2}{c}{\textbf{Assembly101}} \\
        \cmidrule(lr){3-4} \cmidrule(lr){5-6} \cmidrule(lr){7-8}
        & & Top-1/5 Acc & Recall@5 & Top-1/5 Acc & Recall@5 & Top-1/5 Acc & Recall@5 \\
        
        \midrule
        \multirow{2}{*}{None} & RGB & 37.37 / 87.64 & 46.57 & 27.21 / 49.77 &  8.30 &  0.30 /  1.49 &  \underline{4.03} \\
        & RGB, Depth & \underline{38.18 / 86.95} & 48.75 & \underline{27.21 / 50.23} &  8.33 &  0.32 /  1.65 &  3.85 \\

        \midrule
        \multirow{3}{*}{GT} & AH & 63.55 / 94.52 & 65.09 & 31.75 / 73.57 & 37.33 & 31.19 / 59.70 & \underline{38.49} \\
         & RGB, AH & 62.38 / 94.04 & 65.73 & 31.00 / 71.83 & 32.68 & 28.74 / 56.19 & 33.80 \\
         & RGB, Depth, AH & \underline{65.87 / 94.20} & 64.05 & \underline{33.24 / 73.47} & 35.17 & 30.34 / 57.81 & 36.66 \\
        \midrule
        \multirow{3}{*}{Pred} & AH* & 48.02 / 87.56 & 53.58 & 26.43 / 59.81 & 17.32 & 13.14 / 32.28 & \textbf{13.02} \\
        & RGB, AH & 48.90 / 89.32 & 56.34 & 26.85 / 58.11 & 14.39 & 12.57 / 30.20 & 10.97  \\
        & RGB, Depth, AH & \textbf{51.26 / 88.88} & 54.78 & \textbf{27.24 / 60.41} & 16.12 & 12.49 / 30.81 & 12.89 \\
        \bottomrule
    \end{tabular}
\end{table*}

%----------- SubSection--------------------
\subsection{Multimodal Complementarity Analysis}
\noindent To examine how multimodal fusion and temporal context jointly influence anticipation performance under varying levels of dataset complexity and observation uncertainty, we analyze AAG+ from two complementary perspectives. First, we quantify the effect of different modality combinations under both idealized ground-truth and realistic predicted action-history settings. Second, we perform a per-sample analysis of visual and textual contributions to correct and incorrect predictions, revealing conditions under which the modalities act redundantly or complementarily.

%%%--- SUB SUB SECTION-----------
\subsubsection{Effect of Modality Composition under Ground-Truth and Predicted Action Histories}

Table~\ref{tab:modalities} analyzes the contribution of individual and combined modalities in AAG+ under two settings: ground-truth (GT) histories and realistic (Pred) histories populated by an equivalent single-frame action recognizer.

The analysis indicates that when no action history is available, RGB alone provides a reasonable baseline on IKEA-ASM and Meccano, while performing poorly on Assembly101, potentially attributed to the complexity and variability of the tasks in this dataset. Adding depth consistently improves or stabilizes performance across datasets, confirming that geometric cues complement RGB by providing spatial disambiguation, particularly in visually cluttered scenes.

Combining visual cues with action history further improves performance under both GT and predicted action history population settings. Using GT action history yields a substantial performance boost across all datasets, highlighting the importance of long-term procedural context. Under realistic settings, multimodality becomes even more critical. While predicted action history alone is noisy, fusing it with RGB stabilizes predictions, and incorporating depth further enhances robustness, leading to the best performance across all datasets according to their respective evaluation protocols.

From a dataset perspective, it is evident that the impact of multimodality is strongly modulated by dataset complexity. On IKEA-ASM, which features shorter procedural sequences and lower intra-activity variability, strong visual cues are often sufficient for action anticipation, and action history primarily acts as a performance booster. Consequently, gains from multimodal fusion are incremental. In contrast, for Meccano and especially Assembly101 which exhibit longer action sequences, a larger action vocabulary, and higher visual ambiguity at reliance on action history becomes increasingly critical. The textual action history encoder provides a strong procedural prior that helps disambiguate future actions when instantaneous visual evidence is insufficient. Under realistic conditions, multimodal fusion plays a compensatory role: while predicted action histories are imperfect, their integration with RGB and depth stabilizes anticipation by balancing short-term visual evidence with long-term task progression.

%%%---- SUB SUB SECTION---
\subsubsection{Modality Contribution Analysis via Visual-Textual Agreement}

\begin{figure*}
    \centering
    \includegraphics[width=\linewidth]{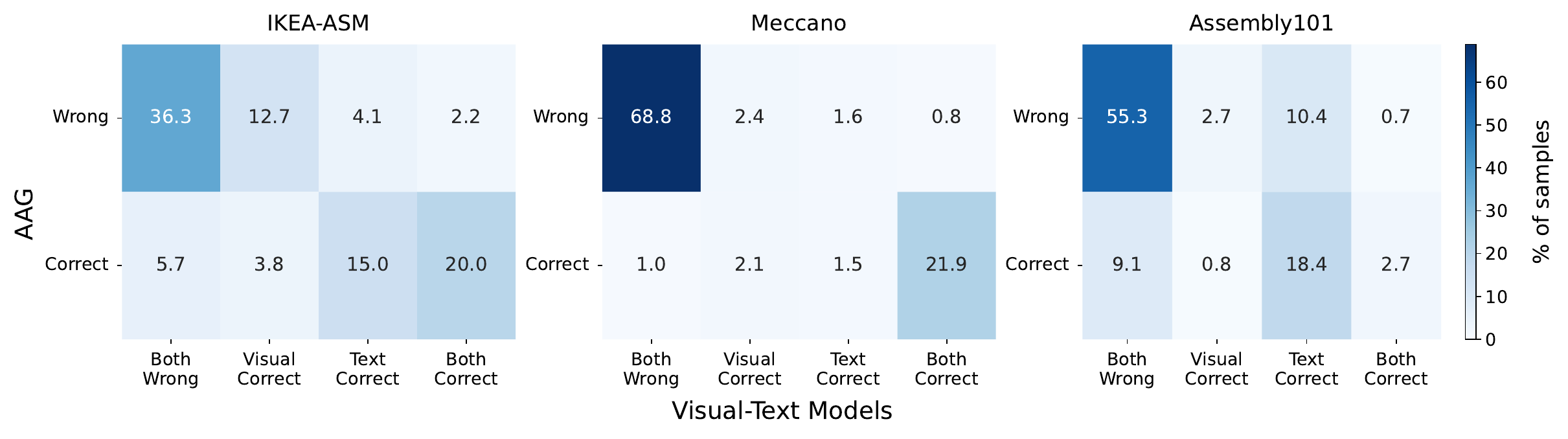}
    \includegraphics[width=\linewidth]{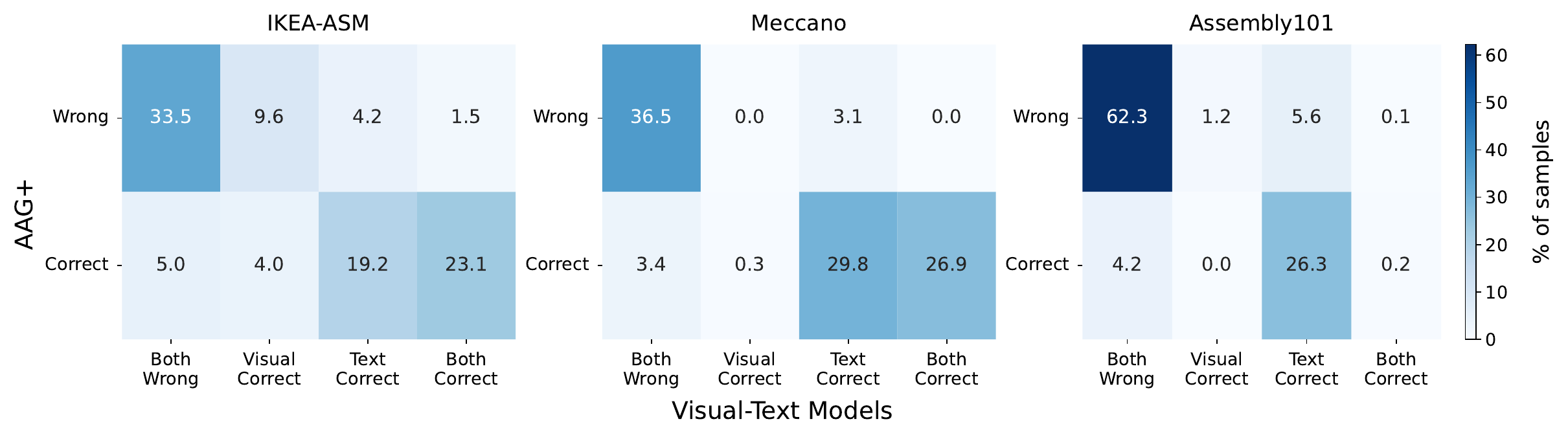}
    \caption{\textbf{Confusion matrix-style modality breakdown for AAG and AAG+ across datasets.} Each sample is categorized by whether the visual branch (RGB-Depth), the text action-history branch, or both produce a correct prediction. }
    \label{fig:Visual-Text}
\end{figure*}

To better understand how different modalities contribute to anticipation performance on each dataset, following~\cite{kontras2025balancing}, we conduct a meta-analysis based on confusion matrix-style breakdowns that quantify, for each sample, whether the visual branch (RGB and depth), the textual action history branch, or both lead to a correct or incorrect prediction. This analysis goes beyond aggregate accuracy and reveals when modalities are complementary versus redundant.

Our analysis, shown in Figure~\ref{fig:Visual-Text}, reveals that while the visual modality provides essential grounding in the physical scene, the text modality often stabilizes predictions, particularly under high visual occlusion or ambiguity, as in Meccano and Assembly101. For both the original AAG and the improved AAG+ models, the text modality contributes most strongly to overall performance, highlighting the importance of long-term procedural context, especially in datasets with many activities, such as IKEA-ASM and Assembly101, and with high intra-activity variation, as in Assembly101.

Comparing AAG and AAG+, both models exhibit dataset-dependent modality preferences. Across all datasets, AAG+ consistently outperforms AAG by shifting probability mass from failure cases toward jointly correct multimodal predictions through its bidirectional cross-attention mechanism. On IKEA-ASM and Meccano, these gains are primarily driven by an increase in jointly correct predictions, indicating improved cross-modal agreement. On more complex datasets, especially Assembly101, AAG+ shows a higher proportion of \textit{Text Correct} cases than AAG, reflecting the increased importance of robustness to imperfect action histories in settings with longer action sequences, higher intra-activity variation, and visually ambiguous intermediate states, where single-frame visual cues are often insufficient. 

We further analyze the contributions of each modality and the effectiveness of the fusion mechanism in Appendix D under visual-only (RGB-Depth) and RGB-Text settings.

%\textcolor{blue}{Comparing AAG and AAG+, both models exhibit dataset-dependent modality preferences. Across all datasets, AAG+ consistently outperforms AAG by reallocating probability mass from failure cases toward joint multimodal correctness thanks to the bidirectional cross-attention mechanism. For IKEA-ASM and Meccano, performance gains are primarily driven by an increase in jointly correct predictions, indicating improved cross-modal agreement. In contrast, Meccano and particularly Assembly101 show a higher proportion of Text Correct predictions in AAG+ relative to AAG. This reflects the increased importance of long-term procedural context in datasets with longer action sequences, greater intra-activity variation, and visually ambiguous intermediate states, where single-frame visual cues are often insufficient. The refinements introduced in AAG+, including more reliable history construction, robustness to imperfect sequences, and improved fusion, enable the model to leverage textual priors when visual evidence is weak, leading to adaptive modality reliance rather than enforced visual dominance.}

\subsection{Impact of Keyframe Selection}
\noindent Table~\ref{tab:keyframe} reports the impact of different keyframe selection strategies across IKEA-ASM, Meccano, and Assembly101 datasets. We compare uniform sampling of the most recent frame (\emph{None}) against distance-based selection using cosine and $L_2$ similarity, as well as blurriness-based selection using the Laplacian variance.

Distance-based strategies are consistently outperformed by the blurriness-based strategy and, in most cases, by not applying any keyframe selection, likely due to noisy embeddings caused by motion blur. In contrast, blurriness-based selection consistently achieves the strongest or comparable performance across metrics. While absolute gains are modest, this approach improves robustness by avoiding visually degraded frames that can negatively affect both RGB and depth feature extraction.

As previously noted, IKEA-ASM and Assembly101 use a threshold of 100, considering frames with Laplacian variance above this value as sharp. As reported in Appendix~A, the mean Laplacian on IKEA-ASM exceeds 300, indicating very sharp frames, so blurriness-based selection yields results comparable to no selection, as the third-person perspective already provides clear, informative frames. Assembly101, on the other hand, has a mean slightly above 100, and discarding frames below this threshold yields measurable improvement.

Finally, Meccano presents the greatest challenge for keyframe selection due to pronounced motion blur from a head-mounted egocentric camera, with a mean Laplacian below 50, indicating pronounced blur. After evaluating thresholds of 30, 50, and 100, we adopt 50, which balances the number of updates while preserving reasonably sharp frames.

Overall, results show that the benefit of keyframe selection depends on the dataset. When most frames are already clear, simply using the most recent frame works well. In settings with more motion blur, explicitly filtering out blurry frames helps improve single-frame anticipation performance.

\begin{table*}[t]
\centering
%\small
\setlength{\tabcolsep}{3pt} % adjust spacing

% Left subtable
\begin{minipage}{0.48\textwidth}
\centering
\captionof{table}{Comparison of keyframe selection strategies.}
\label{tab:keyframe}
\begin{tabular}{lrrrrrr}
\toprule
\textbf{Strategy} & \multicolumn{2}{c}{\textbf{IKEA-ASM}} & \multicolumn{2}{c}{\textbf{Meccano}} & \multicolumn{2}{c}{\textbf{Assembly101}} \\
\cmidrule(lr){2-3} \cmidrule(lr){4-5} \cmidrule(lr){6-7}
 & Acc@1/5 & R@5 & Acc@1/5 & R@5 & Acc@1/5 & R@5 \\ \midrule
None & \textbf{48.9/85.1} & 53.0 & 26.1/56.4 & 20.2 & 12.4/29.9 & 12.1 \\
Cos & 46.2/85.4 & 54.4 & 25.5/59.1 & 20.5 & 12.8/29.7 & 11.7 \\
L2 & 46.4/84.6 & 54.6 & 25.1/57.2 & 18.9 & 12.4/30.7 & 12.4 \\
Blur & \textbf{48.9/85.1} & 53.0 & \textbf{26.4/56.8} & 19.3 & 12.2/30.0 & \textbf{12.6} \\
\bottomrule
\end{tabular}
\end{minipage}
\hfill
% Right subtable
\begin{minipage}{0.48\textwidth}
\centering
\captionof{table}{Comparison of stochastic action history corruption strategies.}
\label{tab:action_swp}
\begin{tabular}{lrrrrrr}
\toprule
\textbf{Strategy} & \multicolumn{2}{c}{\textbf{IKEA-ASM}} & \multicolumn{2}{c}{\textbf{Meccano}} & \multicolumn{2}{c}{\textbf{Assembly101}} \\
\cmidrule(lr){2-3} \cmidrule(lr){4-5} \cmidrule(lr){6-7}
 & Acc@1/5 & R@5 & Acc@1/5 & R@5 & Acc@1/5 & R@5 \\ \midrule
None & 48.9/85.1 & 53.0 & 26.4/56.8 & 19.3 & 12.2/30.0 & 12.6 \\
Noise & 49.8/88.4 & 56.4 & 26.2/58.0 & 16.9 & 12.5/30.8 & \textbf{12.9} \\
Swap $p=0.1$ & \textbf{51.3/88.9} & 54.8 & 24.8/59.1 & 17.0 & 12.2/30.9 & 12.0 \\
Swap $p=0.2$ & 51.2/89.7 & 54.2 & 26.1/57.9 & 17.5 & 12.2/32.0 & 12.5 \\
Swap $p=0.3$ & 50.1/89.9 & 59.1 & \textbf{27.2/60.4} & 16.1 & 12.8/32.4 & 12.1 \\
Swap $p=0.4$ & 48.0/89.2 & 56.9 & 26.9/61.3 & 15.9 & 12.5/33.0 & 12.3 \\
Swap $p=0.5$ & 47.0/87.6 & 53.3 & 26.8/60.7 & 15.9 & 12.9/34.2 & 11.6 \\
\bottomrule
\end{tabular}
\end{minipage}
\label{tab:ablation_summary}
\end{table*}

%%%----- Subsection------------------------
\subsection{Impact of Stochastic Action History Corruptions}
\noindent Table~\ref{tab:action_swp} summarizes the effect of different action history corruption strategies when action histories are populated using predicted actions from equivalent action recognizers, and frames selected using the blurriness-based strategy. %Importantly, we follow the official evaluation protocol of each dataset: performance on IKEA-ASM and Meccano is primarily assessed using Top-1/5 accuracy, whereas for Assembly101 the main indicator is Recall@5, reflecting its larger action space and higher ambiguity.

We observe that performance is strongly influenced by both dataset complexity and the size of the action space. On IKEA-ASM, action swapping with moderate corruption levels ($p=0.1-0.2$) achieve the highest Top-1 accuracy, improving robustness while preserving informative temporal cues. In contrast, Meccano, which has an action space approximately twice as large as IKEA-ASM and relies on a less accurate single-frame action recognizer, benefits from increased variability. Accordingly, higher corruption levels for action swapping ($p=0.3-0.5$) yield the best results.

Assembly101 exhibits a markedly different behavior. Due to its large number of activities ($>1000$) and substantially higher intra- and inter-activity variability, action swapping tends to degrade performance. Instead, applying random embedding noise and dropout provides the strongest gains, suggesting that softer, continuous perturbations are better suited to handling the high ambiguity of this dataset.

\subsection{Impact of Action History Source}
\noindent Table~\ref{tab:ah_source} reports anticipation performance across various action history sources on the IKEA-ASM dataset. This ablation is conducted without action history corruption strategies to isolate the impact of the source itself. We compare histories constructed by a single-frame action recognizer against those generated by GPT-5.1 and Gemini 3 Flash, utilizing varying degrees of contextual information (i.e., short vs. long prompts and sequences, as detailed in Section~\ref{sec:method}). IKEA-ASM was selected for this analysis as its relatively lower task complexity provides a clear baseline for evaluating VLM-based history construction.

Results demonstrate that Gemini 3 Flash outperforms GPT~5.1, while also confirming that accuracy scales with context, a finding consistent with prior literature~\cite{wang2025action,yuan2025video,zou2025unlocking}. Despite this improvement, both VLM-based histories remain significantly inferior to the single-frame action recognizer.

Qualitative analyses in Appendix~E reveal that, while VLMs provide plausible rationales for their predictions, they rely heavily on heuristic priors of common scenarios rather than on precise visual perception. This behavior is consistent with observations in \cite{ahn2025happens} for video question-answering tasks. As a result, action history retrieval is biased toward typical behaviors rather than the specific assembly instructions, which in turn negatively affects anticipation performance, strongly dependent on the accuracy of the action history.

\begin{table}[t]
%\small
\centering
\setlength{\tabcolsep}{3pt}
\caption{Comparison of action history sources.}
\label{tab:ah_source}
\begin{tabular}{lrr}
\toprule
\textbf{Source} & \textbf{Top1/5 Acc} & \textbf{Rec@5} \\ \midrule
Single-Frame Action Recognizer & \textbf{48.90 / 94.20} & \textbf{64.05} \\\midrule
GPT 5.1 (Short Prompt) & 14.37 / 54.14 & 28.18 \\
GPT 5.1 (Long Prompt) & 17.45 / 55.66 & 27.83 \\
GPT 5.1 (Long + Action Seq) & 15.45 / 53.34 & 26.57 \\\midrule
Gemini 3 Flash (Short Prompt) & 19.05 / 68.27 & 36.62 \\
Gemini 3 Flash (Long Prompt) & 20.45 / 67.47 & 35.54 \\
Gemini 3 Flash (Long + Action Seq) & 22.89 / 71.67 & 42.16 \\
\bottomrule
\end{tabular}
\end{table}

\section{Conclusions}
\label{sec:conclusions}

\noindent In this work, we investigated action anticipation under the extreme constraint of single-frame observation, aiming to understand how far performance can be pushed when instantaneous visual cues are combined with structured temporal priors. Building on our previous AAG framework, we introduce AAG+, which incorporates improved keyframe selection, more reliable action-history modeling, robustness to imperfect histories, and more effective multimodal fusion. Our experiments across datasets of increasing complexity show that AAG+ consistently outperforms AAG by shifting predictions away from failure cases toward joint visual–textual correctness.

A central finding is that, when equipped with state-of-the-art single-frame foundation models for appearance and geometry, single-frame anticipation can approach, and in some cases surpass, the performance of methods that explicitly process full video clips. This indicates that much of the information exploited by video-based models is implicitly captured by strong per-frame representations, provided that long-term procedural context is explicitly encoded and leveraged. The relative contribution of modalities is strongly dataset-dependent: visual cues dominate in short, visually unambiguous procedures, while textual action-history representations gain importance in long-horizon, high-variability settings. Crucially, AAG+ learns to adapt its reliance on visual versus textual cues rather than enforcing a fixed modality hierarchy.

Beyond performance gains, this work offers empirical insight into the limits and capabilities of single-frame anticipation. It demonstrates that, with modern foundation models and principled multimodal fusion, single-frame approaches provide a competitive and computationally efficient alternative to video-based methods, particularly in low-latency settings where multi-frame processing is impractical. These findings offer a foundation for future research on anticipation, procedural reasoning, and decision-making under partial observability.

\section*{Acknowledgments}
\noindent This work has been funded by the Spanish State Research Agency (AEI) and ERDF/EU under grant: GEMELIA PID2024-161711OB-I00. This work has also been supported by a Spanish national fund for PhD studies, (FPU21/00414) as well as by the European Union (EU - HE Magician – Grant Agreement 101120731).

{
    \small
    \bibliographystyle{IEEEtran}
    \bibliography{main}

@String(CVPR= {IEEE Conf. Comput. Vis. Pattern Recog.})

@String(ICCV= {Int. Conf. Comput. Vis.})

@String(ECCV= {Eur. Conf. Comput. Vis.})

@String(BMVC= {Brit. Mach. Vis. Conf.})

@String(ICIP = {IEEE Int. Conf. Image Process.})

@String(ICLR = {Int. Conf. Learn. Represent.})

@String(CVPR  = {CVPR})

@String(ICCV  = {ICCV})

@String(ECCV  = {ECCV})

@String(BMVC  =	{BMVC})

@String(ICIP  = {ICIP})

@String(ICLR  = {ICLR})

@article{rasouli2020pedestrianactionanticipationusing,
  title={Pedestrian Action Anticipation using Contextual Feature Fusion in Stacked RNNs}, 
  author={Amir Rasouli and Iuliia Kotseruba and John K. Tsotsos},
  year={2020},
  journal={BMVC}
}

@article{liu2020spatiotemporal,
  title={Spatiotemporal relationship reasoning for pedestrian intent prediction},
  author={Liu, Bingbin and Adeli, Ehsan and Cao, Zhangjie and Lee, Kuan-Hui and Shenoi, Abhijeet and Gaidon, Adrien and Niebles, Juan Carlos},
  journal={IEEE Robotics and Automation Letters},
  volume={5},
  number={2},
  pages={3485--3492},
  year={2020},
  publisher={IEEE}
}

@inproceedings{girase2021loki,
  title={Loki: Long term and key intentions for trajectory prediction},
  author={Girase, Harshayu and Gang, Haiming and Malla, Srikanth and Li, Jiachen and Kanehara, Akira and Mangalam, Karttikeya and Choi, Chiho},
  booktitle={Proceedings of the IEEE/CVF International Conference on Computer Vision},
  pages={9803--9812},
  year={2021}
}

@InProceedings{Sener_2022_CVPR,
    author    = {Sener, Fadime and Chatterjee, Dibyadip and Shelepov, Daniel and He, Kun and Singhania, Dipika and Wang, Robert and Yao, Angela},
    title     = {Assembly101: A Large-Scale Multi-View Video Dataset for Understanding Procedural Activities},
    booktitle = {Proceedings of the IEEE/CVF Conference on Computer Vision and Pattern Recognition (CVPR)},
    month     = {June},
    year      = {2022},
    pages     = {21096-21106}
}

@InProceedings{Ben-Shabat_2021_WACV,
    author    = {Ben-Shabat, Yizhak and Yu, Xin and Saleh, Fatemeh and Campbell, Dylan and Rodriguez-Opazo, Cristian and Li, Hongdong and Gould, Stephen},
    title     = {The IKEA ASM Dataset: Understanding People Assembling Furniture Through Actions, Objects and Pose},
    booktitle = {Proceedings of the IEEE/CVF Winter Conference on Applications of Computer Vision (WACV)},
    month     = {January},
    year      = {2021},
    pages     = {847-859}
}

@InProceedings{Manousaki_2023_ICCV,
    author    = {Manousaki, Victoria and Bacharidis, Konstantinos and Papoutsakis, Konstantinos and Argyros, Antonis},
    title     = {VLMAH: Visual-Linguistic Modeling of Action History for Effective Action Anticipation},
    booktitle = {Proceedings of the IEEE/CVF International Conference on Computer Vision (ICCV) Workshops},
    month     = {October},
    year      = {2023},
    pages     = {1917-1927}
}

@article{benaventlledo2024enhancing,
title = {Enhancing action recognition by leveraging the hierarchical structure of actions and textual context},
journal = {Computer Vision and Image Understanding},
volume = {262},
pages = {104560},
year = {2025},
issn = {1077-3142},
doi = {10.1016/j.cviu.2025.104560},
author = {Manuel Benavent-Lledo and David Mulero-Pérez and David Ortiz-Perez and Jose Garcia-Rodriguez and Antonis Argyros},
}

@article{zhao2023antgpt,
        title={AntGPT: Can Large Language Models Help Long-term Action Anticipation from Videos?},
        author={Qi Zhao and Shijie Wang and Ce Zhang and Changcheng Fu and Minh Quan Do and Nakul Agarwal and Kwonjoon Lee and Chen Sun},
        journal={ICLR},
        year={2024}
     }

@article{RAGUSA2023103764,
title = {MECCANO: A multimodal egocentric dataset for humans behavior understanding in the industrial-like domain},
journal = {Computer Vision and Image Understanding},
volume = {235},
pages = {103764},
year = {2023},
issn = {1077-3142},
doi = {https://doi.org/10.1016/j.cviu.2023.103764},
author = {Francesco Ragusa and Antonino Furnari and Giovanni Maria Farinella}}

@inproceedings{NEURIPS2024_26cfdcd8,
 author = {Yang, Lihe and Kang, Bingyi and Huang, Zilong and Zhao, Zhen and Xu, Xiaogang and Feng, Jiashi and Zhao, Hengshuang},
 booktitle = {Advances in Neural Information Processing Systems},
 editor = {A. Globerson and L. Mackey and D. Belgrave and A. Fan and U. Paquet and J. Tomczak and C. Zhang},
 pages = {21875--21911},
 publisher = {Curran Associates, Inc.},
 title = {Depth Anything V2},
 volume = {37},
 year = {2024}
}

@inproceedings{lin2019tsm,
  title={Tsm: Temporal shift module for efficient video understanding},
  author={Lin, Ji and Gan, Chuang and Han, Song},
  booktitle={Proceedings of the IEEE/CVF International Conference on Computer Vision},
  pages={7083--7093},
  year={2019}
}

@inproceedings{lan2014hierarchical,
  title={A hierarchical representation for future action prediction},
  author={Lan, Tian and Chen, Tsung-Chuan and Savarese, Silvio},
  booktitle={Computer Vision--ECCV 2014: 13th European Conference, Zurich, Switzerland, September 6-12, 2014, Proceedings, Part III 13},
  pages={689--704},
  year={2014},
  organization={Springer}
}

@inproceedings{vondrick2016anticipating,
  title={Anticipating visual representations from unlabeled video},
  author={Vondrick, Carl and Pirsiavash, Hamed and Torralba, Antonio},
  booktitle={Proceedings of the IEEE conference on computer vision and pattern recognition},
  pages={98--106},
  year={2016}
}

@inproceedings{AVT,
  title={Anticipative video transformer},
  author={Girdhar, Rohit and Grauman, Kristen},
  booktitle={Proceedings of the IEEE/CVF international conference on computer vision},
  pages={13505--13515},
  year={2021}
}

@inproceedings{TempAgg,
  title={Temporal aggregate representations for long-range video understanding},
  author={Sener, Fadime and Singhania, Dipika and Yao, Angela},
  booktitle={Computer Vision--ECCV 2020: 16th European Conference, Glasgow, UK, August 23--28, 2020, Proceedings, Part XVI 16},
  pages={154--171},
  year={2020},
  organization={Springer}
}

@article{RULSTM,
  title={Rolling-unrolling lstms for action anticipation from first-person video},
  author={Furnari, Antonino and Farinella, Giovanni Maria},
  journal={IEEE transactions on pattern analysis and machine intelligence},
  volume={43},
  number={11},
  pages={4021--4036},
  year={2020},
  publisher={IEEE}
}

@article{sampat2024actioncomet,
  title={ActionCOMET: A Zero-shot Approach to Learn Image-specific Commonsense Concepts about Actions},
  author={Sampat, Shailaja Keyur and Yang, Yezhou and Baral, Chitta},
  journal={arXiv preprint arXiv:2410.13662},
  year={2024}
}

@inproceedings{park2020visualcomet,
  title={Visualcomet: Reasoning about the dynamic context of a still image},
  author={Park, Jae Sung and Bhagavatula, Chandra and Mottaghi, Roozbeh and Farhadi, Ali and Choi, Yejin},
  booktitle={Computer Vision--ECCV 2020: 16th European Conference, Glasgow, UK, August 23--28, 2020, Proceedings, Part V 16},
  pages={508--524},
  year={2020},
  organization={Springer}
}

@inproceedings{jang2020etri,
  title={ETRI-activity3D: A large-scale RGB-D dataset for robots to recognize daily activities of the elderly},
  author={Jang, Jinhyeok and Kim, Dohyung and Park, Cheonshu and Jang, Minsu and Lee, Jaeyeon and Kim, Jaehong},
  booktitle={2020 IEEE/RSJ International Conference on Intelligent Robots and Systems (IROS)},
  pages={10990--10997},
  year={2020},
  organization={IEEE}
}

@article{hwang2021eldersim,
  title={Eldersim: A synthetic data generation platform for human action recognition in eldercare applications},
  author={Hwang, Hochul and Jang, Cheongjae and Park, Geonwoo and Cho, Junghyun and Kim, Ig-Jae},
  journal={IEEE Access},
  volume={11},
  pages={9279--9294},
  year={2021},
  publisher={IEEE}
}

@article{dai2022toyota,
  title={Toyota smarthome untrimmed: Real-world untrimmed videos for activity detection},
  author={Dai, Rui and Das, Srijan and Sharma, Saurav and Minciullo, Luca and Garattoni, Lorenzo and Bremond, Francois and Francesca, Gianpiero},
  journal={IEEE Transactions on Pattern Analysis and Machine Intelligence},
  volume={45},
  number={2},
  pages={2533--2550},
  year={2022},
  publisher={IEEE}
}

@article{lai2024human,
  title={Human Action Anticipation: A Survey},
  author={Lai, Bolin and Toyer, Sam and Nagarajan, Tushar and Girdhar, Rohit and Zha, Shengxin and Rehg, James M and Kitani, Kris and Grauman, Kristen and Desai, Ruta and Liu, Miao},
  journal={arXiv preprint arXiv:2410.14045},
  year={2024}
}

@article{oquab2023dinov2,
  title={Dinov2: Learning robust visual features without supervision},
  author={Oquab, Maxime and Darcet, Timoth{\'e}e and Moutakanni, Th{\'e}o and Vo, Huy and Szafraniec, Marc and Khalidov, Vasil and Fernandez, Pierre and Haziza, Daniel and Massa, Francisco and El-Nouby, Alaaeldin and others},
  journal={arXiv preprint arXiv:2304.07193},
  year={2023}
}

@article{sanh2019distilbert,
  title={DistilBERT, a distilled version of BERT: smaller, faster, cheaper and lighter},
  author={Sanh, Victor and Debut, Lysandre and Chaumond, Julien and Wolf, Thomas},
  journal={arXiv preprint arXiv:1910.01108},
  year={2019}
}

@article{damen2022rescaling,
  title={Rescaling egocentric vision: Collection, pipeline and challenges for epic-kitchens-100},
  author={Damen, Dima and Doughty, Hazel and Farinella, Giovanni Maria and Furnari, Antonino and Kazakos, Evangelos and Ma, Jian and Moltisanti, Davide and Munro, Jonathan and Perrett, Toby and Price, Will and others},
  journal={International Journal of Computer Vision},
  pages={1--23},
  year={2022},
  publisher={Springer}
}

@article{lei2022revealing,
  title={Revealing single frame bias for video-and-language learning},
  author={Lei, Jie and Berg, Tamara L and Bansal, Mohit},
  journal={arXiv preprint arXiv:2206.03428},
  year={2022}
}

@inproceedings{zatsarynna2023action,
  title={Action anticipation with goal consistency},
  author={Zatsarynna, Olga and Gall, Juergen},
  booktitle={2023 IEEE International Conference on Image Processing (ICIP)},
  pages={1630--1634},
  year={2023},
  organization={IEEE}
}

@INPROCEEDINGS{plausiVL2024,
  author={Mittal, Himangi and Agarwal, Nakul and Lo, Shao-Yuan and Lee, Kwonjoon},
  booktitle={2024 IEEE/CVF Conference on Computer Vision and Pattern Recognition (CVPR)}, 
  title={Can't make an Omelette without Breaking some Eggs: Plausible Action Anticipation using Large Video-Language Models}, 
  year={2024},
  volume={},
  number={},
  pages={18580-18590},
  doi={10.1109/CVPR52733.2024.01758}}

@article{cao2025vision,
  title={Vision and Intention Boost Large Language Model in Long-Term Action Anticipation},
  author={Cao, Congqi and Hu, Lanshu and Yu, Yating and Zhang, Yanning},
  journal={arXiv preprint arXiv:2505.01713},
  year={2025}
}

@inproceedings{sato2025bidirectional,
  title={Bidirectional Action Sequence Learning for Long-term Action Anticipation with Large Language Models},
  author={Sato, Yuji and Ishii, Yasunori and Yamashita, Takayoshi},
  booktitle={2025 19th International Conference on Machine Vision and Applications (MVA)},
  pages={1--6},
  year={2025},
  organization={IEEE}
}

@InProceedings{10.1007/978-3-319-10602-1_28,
author="Vu, Tuan-Hung
and Olsson, Catherine
and Laptev, Ivan
and Oliva, Aude
and Sivic, Josef",
editor="Fleet, David
and Pajdla, Tomas
and Schiele, Bernt
and Tuytelaars, Tinne",
title="Predicting Actions from Static Scenes",
booktitle="Computer Vision -- ECCV 2014",
year="2014",
publisher="Springer International Publishing",
address="Cham",
pages="421--436",
isbn="978-3-319-10602-1"
}

@article{wang2025actionllm,
  title={Multimodal Large Models Are Effective Action Anticipators},
  author={Wang, Binglu and Tian, Yao and Wang, Shunzhou and Yang, Le},
  journal={IEEE Transactions on Multimedia},
  year={2025},
  publisher={IEEE}
}

@inproceedings{gkioxari2015contextual,
  title={Contextual action recognition with r* cnn},
  author={Gkioxari, Georgia and Girshick, Ross and Malik, Jitendra},
  booktitle={Proceedings of the IEEE international conference on computer vision},
  pages={1080--1088},
  year={2015}
}

@article{liang2024patch,
  title={Patch excitation network for boxless action recognition in still images},
  author={Liang, Shuang and Wang, Jiewen and Zhuang, Zikun},
  journal={The Visual Computer},
  volume={40},
  number={6},
  pages={4099--4113},
  year={2024},
  publisher={Springer}
}

@article{ashrafi2023still,
  title={Still image action recognition based on interactions between joints and objects},
  author={Ashrafi, Seyed Sajad and Shokouhi, Shahriar B and Ayatollahi, Ahmad},
  journal={Multimedia Tools and Applications},
  volume={82},
  number={17},
  pages={25945--25971},
  year={2023},
  publisher={Springer}
}

@inproceedings{hosseyni2024human,
  title={Human Action Recognition in Still Images Using ConViT},
  author={Hosseyni, Seyed Rohollah and Seyedin, Sanaz and Taheri, Hassan},
  booktitle={2024 32nd International Conference on Electrical Engineering (ICEE)},
  pages={1--7},
  year={2024},
  organization={IEEE}
}

@inproceedings{saleknia2024multi,
  title={Multi step knowledge distillation framework for action recognition in still images},
  author={Saleknia, Amir Hossein and Ayatollahi, Ahmad},
  booktitle={2024 20th CSI International Symposium on Artificial Intelligence and Signal Processing (AISP)},
  pages={1--7},
  year={2024},
  organization={IEEE}
}

@inproceedings{he2023context,
  title={Context Enhancement Methodology for Action Recognition in Still Images},
  author={He, Jiarong and Wu, Wei and Li, Yuxing},
  booktitle={International Conference on Artificial Neural Networks},
  pages={112--122},
  year={2023},
  organization={Springer}
}

@InProceedings{Feichtenhofer_2019_ICCV,
author = {Feichtenhofer, Christoph and Fan, Haoqi and Malik, Jitendra and He, Kaiming},
title = {SlowFast Networks for Video Recognition},
booktitle = {Proceedings of the IEEE/CVF International Conference on Computer Vision (ICCV)},
month = {October},
year = {2019}
}

@inproceedings{carreira2017quo,
  title={Quo vadis, action recognition? a new model and the kinetics dataset},
  author={Carreira, Joao and Zisserman, Andrew},
  booktitle={proceedings of the IEEE Conference on Computer Vision and Pattern Recognition},
  pages={6299--6308},
  year={2017}
}

@inproceedings{aag,
  title={Action Anticipation at a Glimpse: To What Extent Can Multimodal Cues Replace Video?},
  author={Benavent-Lledo, Manuel and Bacharidis, Konstantinos and Manousaki, Victoria and Papoutsakis, Konstantinos and Argyros, Antonis and Garcia-Rodriguez, Jose},
  booktitle={Proceedings of the IEEE/CVF Winter Conference on Applications of Computer Vision (WACV) 2026},
  year={2026}
}

@article{ahn2025happens,
  title={What Happens When: Learning Temporal Orders of Events in Videos},
  author={Ahn, Daechul and Choi, Yura and Choi, Hyeonbeom and Cho, Seongwon and Kim, San and Choi, Jonghyun},
  journal={arXiv preprint arXiv:2512.08979},
  year={2025}
}

@article{sun2025frames,
  title={From Frames to Clips: Training-free Adaptive Key Clip Selection for Long-Form Video Understanding},
  author={Sun, Guangyu and Singhal, Archit and Uzkent, Burak and Shah, Mubarak and Chen, Chen and Kessler, Garin},
  journal={arXiv preprint arXiv:2510.02262},
  year={2025}
}

@article{liang2024keyvideollm,
  title={Keyvideollm: Towards large-scale video keyframe selection},
  author={Liang, Hao and Li, Jiapeng and Bai, Tianyi and Huang, Xijie and Sun, Linzhuang and Wang, Zhengren and He, Conghui and Cui, Bin and Chen, Chong and Zhang, Wentao},
  journal={arXiv preprint arXiv:2407.03104},
  year={2024}
}

@article{team2023gemini,
  title={Gemini: a family of highly capable multimodal models},
  author={Team, Gemini and others},
  journal={arXiv preprint arXiv:2312.11805},
  year={2023}
}

@article{singh2025openai,
  title={OpenAI GPT-5 System Card},
  author={Singh, Aaditya and others},
  journal={arXiv preprint arXiv:2601.03267},
  year={2025}
}

@inproceedings{kontras2025balancing,
  title={Balancing multimodal training through game-theoretic regularization},
  author={Kontras, Konstantinos and Strypsteen, Thomas and Chatzichristos, Christos and Liang, Paul Pu and Blaschko, Matthew B and De Vos, Maarten},
  booktitle={The Thirty-ninth Annual Conference on Neural Information Processing Systems},
  year={2025}
}

@article{yang2025video,
  title={Video Finetuning Improves Reasoning Between Frames},
  author={Yang, Ruiqi and Yun, Tian and Wang, Zihan and Pavlick, Ellie},
  journal={arXiv preprint arXiv:2511.12868},
  year={2025}
}

@article{doorenbos2025video,
  title={Video Panels for Long Video Understanding},
  author={Doorenbos, Lars and Spurio, Federico and Gall, Juergen},
  journal={arXiv preprint arXiv:2509.23724},
  year={2025}
}

@inproceedings{zhou2025glimpse,
  title={Glimpse: Do large vision-language models truly think with videos or just glimpse at them?},
  author={Zhou, Yiyang and Li, Linjie and Qiu, Shi and Yang, Zhengyuan and Zhao, Yuyang and Han, Siwei and He, Yangfan and Li, Kangqi and Ji, Haonian and Zhao, Zihao and others},
  booktitle={Proceedings of the 2025 Conference on Empirical Methods in Natural Language Processing},
  pages={27830--27844},
  year={2025}
}

@article{guo2025towards,
  title={Towards Object-centric Understanding for Instructional Videos},
  author={Guo, Wenliang and Kong, Yu},
  journal={arXiv preprint arXiv:2512.03479},
  year={2025}
}

@inproceedings{zhang2024object,
  title={Object-centric video representation for long-term action anticipation},
  author={Zhang, Ce and Fu, Changcheng and Wang, Shijie and Agarwal, Nakul and Lee, Kwonjoon and Choi, Chiho and Sun, Chen},
  booktitle={Proceedings of the IEEE/CVF Winter Conference on Applications of Computer Vision},
  pages={6751--6761},
  year={2024}
}

@article{kodathala2025temporal,
  title={Temporal vs. Spatial: Comparing DINOv3 and V-JEPA2 Feature Representations for Video Action Analysis},
  author={Kodathala, Sai Varun and Vunnam, Rakesh},
  journal={arXiv preprint arXiv:2509.21595},
  year={2025}
}

@inproceedings{wang2025action,
  title={Action Detail Matters: Refining Video Recognition with Local Action Queries},
  author={Wang, Mengmeng and Huang, Zeyi and Kong, Xiangjie and Shen, Guojiang and Dai, Guang and Wang, Jingdong and Liu, Yong},
  booktitle={Proceedings of the Computer Vision and Pattern Recognition Conference},
  pages={19132--19142},
  year={2025}
}

@article{yuan2025video,
  title={Video-star: Reinforcing open-vocabulary action recognition with tools},
  author={Yuan, Zhenlong and Qu, Xiangyan and Qian, Chengxuan and Chen, Rui and Tang, Jing and Sun, Lei and Chu, Xiangxiang and Zhang, Dapeng and Wang, Yiwei and Cai, Yujun and others},
  journal={arXiv preprint arXiv:2510.08480},
  year={2025}
}

@article{zou2025unlocking,
  title={Unlocking Vision-Language Models for Video Anomaly Detection via Fine-Grained Prompting},
  author={Zou, Shu and Tian, Xinyu and Wesemann, Lukas and Waschkowski, Fabian and Yang, Zhaoyuan and Zhang, Jing},
  journal={arXiv preprint arXiv:2510.02155},
  year={2025}
}

@article{simeoni2025dinov3,
  title={Dinov3},
  author={Sim{\'e}oni, Oriane and Vo, Huy V and Seitzer, Maximilian and Baldassarre, Federico and Oquab, Maxime and Jose, Cijo and Khalidov, Vasil and Szafraniec, Marc and Yi, Seungeun and Ramamonjisoa, Micha{\"e}l and others},
  journal={arXiv preprint arXiv:2508.10104},
  year={2025}
}

@article{lin2025depth,
  title={Depth anything 3: Recovering the visual space from any views},
  author={Lin, Haotong and Chen, Sili and Liew, Junhao and Chen, Donny Y and Li, Zhenyu and Shi, Guang and Feng, Jiashi and Kang, Bingyi},
  journal={arXiv preprint arXiv:2511.10647},
  year={2025}
}

@ARTICLE{10123038,
  author={Xu, Peng and Zhu, Xiatian and Clifton, David A.},
  journal={IEEE Transactions on Pattern Analysis and Machine Intelligence}, 
  title={Multimodal Learning With Transformers: A Survey}, 
  year={2023},
  volume={45},
  number={10},
  pages={12113-12132},
  doi={10.1109/TPAMI.2023.3275156}}
}

\begin{IEEEbiography}[{\includegraphics[width=1in,height=1.25in,clip,keepaspectratio]{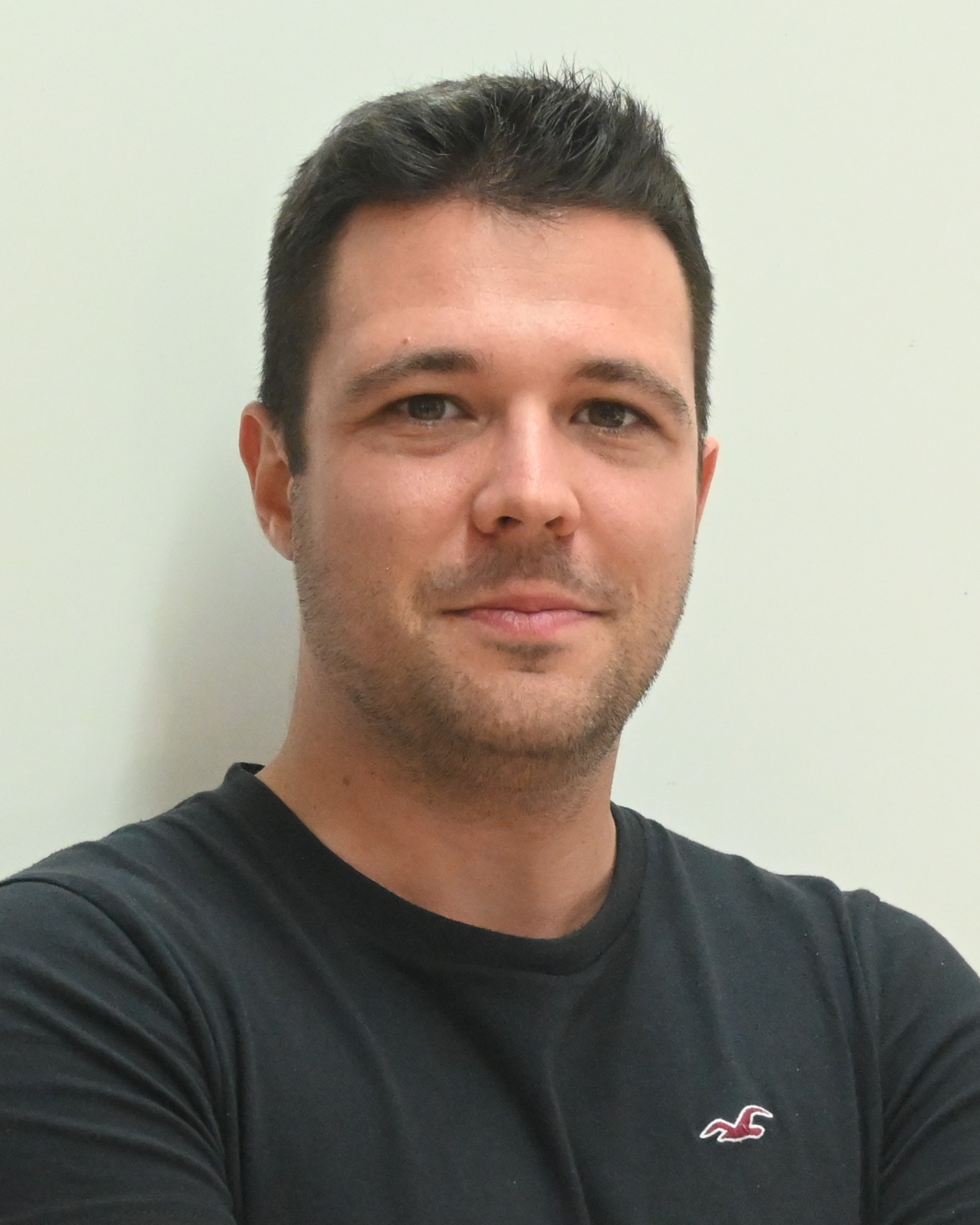}}]{Manuel Benavent-Lledo} received his B.Sc. in Computer Science and M.Sc in Automation and Robotics from the University of Alicante, Spain (2021, 2022), and his PhD in deep learning and computer vision from the University of Alicante in 2025. He is currently a postdoctoral researcher at the Department of Computer Technology, University of Alicante. His research interests include computer vision and deep learning for human action understanding and multimodal fusion. 
\end{IEEEbiography}

\begin{IEEEbiography}[{\includegraphics[width=1in,height=1.25in,clip,keepaspectratio]{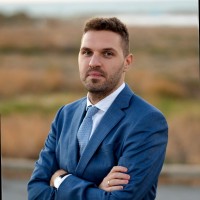}}]{Konstantinos Bacharidis} received his B.Sc. and M.Sc. from the School of Electrical and Computer Engineering, Technical University of Crete (2014, 2016), and his Ph.D. in Computer Science from the University of Crete (2024), where he is a postdoctoral researcher. He is affiliated with the Human-Centered Computer Vision group at the Computational Vision and Robotics Laboratory, FORTH. His research focuses on human action recognition, anticipation, and error detection.
\end{IEEEbiography}

\begin{IEEEbiography}[{\includegraphics[width=1in,height=1.25in,clip,keepaspectratio]{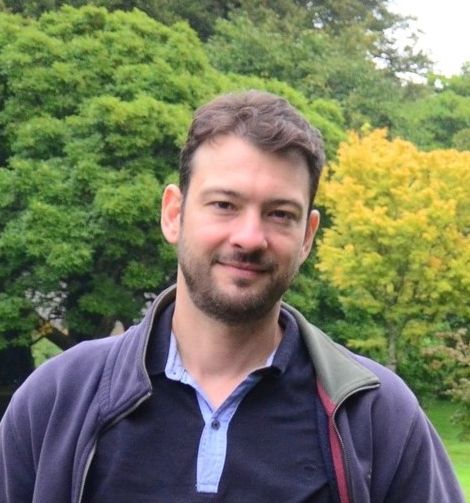}}]{Konstantinos Papoutsakis} is a postdoctoral researcher at the Institute of Computer Science, FORTH, in Heraklion, Crete, Greece. He earned his diploma in Computer Engineering and Informatics from the University of Patras in Greece in 2007, his M.Sc. and PhD degrees in Computer Science from the University of Crete in 2010 and 2019, respectively. His main research interests fall in the areas of computer vision, machine learning and visual perception for robotics with emphasis on human motion analysis, video segmentation, action segmentation, and recognition, and human-computer and human-robot interaction. He has also been actively involved in several EU-funded projects as a computer vision engineer during his postgraduate studies. He has served as PI for a two-year research project (InterLinK) funded by the Hellenic Foundation for Research and Innovation (HFRI).
\end{IEEEbiography}

\begin{IEEEbiography}[{\includegraphics[width=1in,height=1.25in,clip,keepaspectratio]{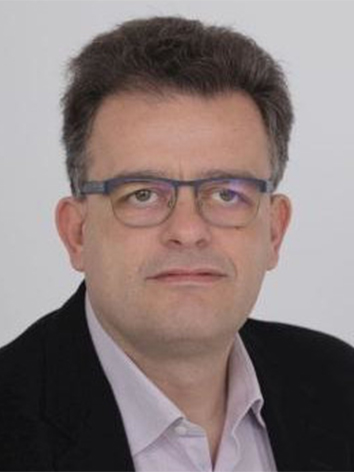}}]{Antonis Argyros} is a professor of computer science at the Computer Science Department, University of Crete and a researcher at the Institute of Computer Science, FORTH, in Heraklion, Crete, Greece. His current research interests fall in the areas of computer vision and pattern recognition, with emphasis on human body pose and shape analysis and on the recognition of human gestures, actions, activities and intentions. He is also interested in applications of computer vision in the fields of robotics and smart environments.
\end{IEEEbiography}

\begin{IEEEbiography}[{\includegraphics[width=1in,height=1.25in,clip,keepaspectratio]{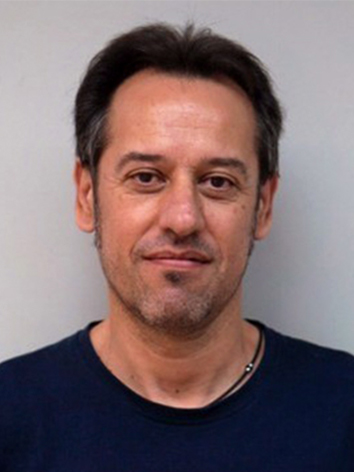}}]{Jose Garcia-Rodriguez} received his Ph.D. degree, with specialization in Computer Vision and Neural Networks, from the University of Alicante (Spain). He is currently Full Professor at the Department of Computer Technology of the University of Alicante. His research areas of interest include: computer vision, computational intelligence, machine learning, pattern recognition, robotics, man-machine interfaces, ambient intelligence, computational chemistry, and parallel and multicore architectures.
\end{IEEEbiography}

%\begin{IEEEbiographynophoto}{John Doe}
%Use $\backslash${\tt{begin\{IEEEbiographynophoto\}}} and the author name as the argument followed by the biography text.
%\end{IEEEbiographynophoto}

\vfill

%\begin{comment}
    
\clearpage
\appendices
\setcounter{section}{0}
\setcounter{table}{0}
\setcounter{figure}{0}
\setcounter{page}{1}
\pagenumbering{roman}
\renewcommand{\thefigure}{S.\arabic{figure}}
\renewcommand{\thetable}{S.\Roman{table}}

\begin{table*}
\begin{center}
    {\Huge \normalfont \papertitle}\\[1em]
    {\Large \normalfont Supplementary Material}
\end{center}
\end{table*}
%\textcolor{red}{Note: All supplemental material must be submitted as separate files and must not be included within the same PDF file as the main paper submission. There is no page limit on supplemental files.}

\noindent The supplementary material is organized as follows. Appendix~\ref{app:ablation} presents a comprehensive ablation study on the impact of different fusion strategies and feature extractors, as well as an analysis of blurriness per dataset. Appendix~\ref{app:ar_results} reports action recognition results for each dataset under different input features and compares them with existing baselines. Appendix~\ref{app:comp_analysis} provides a comparison of AAG+ with state-of-the-art methods in terms of computational efficiency. Finally, Appendix~\ref{app:vlms} presents qualitative results for prompting strategies on vision-language models under the different strategies described in the main paper.

\section{Ablation Experiments}\label{app:ablation}
\noindent As discussed in the main paper, our goal is to enhance the method proposed in \cite{aag} by improving both the unimodal representations (RGB, Depth, Text) and their multimodal integration. To this end, we perform ablation studies on the IKEA-ASM dataset, analyzing different multimodal fusion strategies and evaluating the effectiveness of various feature extractors.

\subsection{Multimodal Fusion}
\noindent The original analysis in \cite{aag} showed that fusing RGB and depth through a cross-attention mechanism, where RGB queries attend to depth, is an effective way to inject geometric cues into the visual representation. This geometry-enhanced RGB embedding outperformed simpler fusion strategies such as concatenation, summation, mean fusion, or self-attention. The same paper also explored visual-textual fusion strategies based on simple operators, including concatenation, summation, mean fusion, and self-attention over visual and textual tokens. Among these, self-attention was the most effective. However, the multimodal analysis shows that it still fails in several multimodal cases: the model produces correct predictions when using visual or textual features independently, yet degrades when both modalities are combined. This inconsistency motivates our exploration of more expressive, modality-aware fusion mechanisms. Results are presented in Table~\ref{tab:ablation_fusion}. Note that we evaluate under predicted action histories only those variants that showed the most promising performance under the perfect (GT) setting and that the visual feature is DINOv2~\cite{oquab2023dinov2}, as in the original AAG framework.

\vspace{0.1cm}
\noindent \textbf{Self-Attention Baseline.} 
We begin by revisiting the self-attention fusion module used in AAG, which serves as the baseline for our analysis. Let $X_v$ and $X_t$ denote the visual and textual feature sequences, respectively. The concatenated feature sequence is then
\begin{equation}
X = [\,X_v \,;\, X_t\,].
\end{equation}
The module computes queries, keys, and values via learned projections:
\begin{equation}
    Q = W_Q X, \qquad
    K = W_K X, \qquad
    V = W_V X.
    \label{eq:qkc_suppl}
\end{equation}
The fused representation is then obtained through the standard attention operator:
\begin{equation}
    X'
    =
    \text{softmax}\!\left(
        \frac{Q K^\top}{\sqrt{D}}
    \right) V,
    \label{eq:attention_suppl}
\end{equation}

where $D$ denotes the feature dimensionality. Although effective in most cases, this mechanism does not adequately control all interactions across modalities and therefore exhibits limited fusion behavior in the visual-textual setting.

\vspace{0.1cm}
\noindent \textbf{Cross-Attention.}
To introduce directional, modality-aware interactions, we compute the queries, keys, and values for cross-modal fusion as:
\begin{equation}
    Q = W_Q X_\text{src}, \qquad
    K = W_K X_\text{tgt}, \qquad
    V = W_V X_\text{tgt},
    \label{eq:qkc_suppl_ca}
\end{equation}
where $X_\text{src}$ and $X_\text{tgt}$ denote the source and target modalities (visual or textual). This asymmetric formulation allows one modality to attend to the other. Throughout this section, we refer to each configuration by the source of the queries, e.g., $\text{CA} (Q = X_t)$ or $\text{CA} (Q = X_v)$.

Results in the second block of Table~\ref{tab:ablation_fusion} show that using the visual modality as source is significantly less effective under ground-truth action histories. In contrast, the configuration $\text{CA} (Q = X_t)$ yields much stronger performance, confirming that complementing textual representations with visual cues is the more informative direction of fusion. However, this advantage diminishes under predicted action histories, where errors from the single-frame action recognizer propagate into the textual embeddings, causing a noticeable drop in accuracy.

To mitigate this sensitivity, we further refine the cross-attention module by introducing \emph{residual cross-attention}, which integrates the source modality with the attended representation. Formally, given the attended output $X'$ (as defined in Equation~\ref{eq:attention_suppl}) and the source modality $X_\text{src}$, we evaluate several fusion operators of the form:
\begin{equation}
Y = f(X_\text{src}, X'),
\end{equation}
including additive residuals, mean fusion, element-wise products, and gated combinations.  
Among these, the gated variant introduces a learnable interpolation between the original and the attended representation:
\begin{equation}
    Y = g \odot X_\text{src} + (1 - g) \odot X',
\end{equation}
where $g \in [0,1]$ is a learned scalar or vector gate.  

These mechanisms aim to stabilize the fusion process by preserving modality-specific information while still benefiting from cross-modal interaction. Nonetheless, the lack of information independent of the action recognizer accuracy hinders the performance of the model under a realistic setting, with the self-attention baseline providing stronger results than the evaluated approaches.

\begin{table*}[t]

\begin{minipage}{0.55\textwidth}
\centering
\setlength{\tabcolsep}{3pt}
\caption{Comparison of multimodal fusion strategies on IKEA-ASM.\\\footnotesize
SA = Self-Attention;
CA = Cross-Attention; 
Bi-CA = Bidirectional Cross-Attention.}
\label{tab:ablation_fusion}
\begin{tabular}{lrr|rr}
\toprule
\multirow{2}{*}{Fusion Strategy} & \multicolumn{2}{c|}{GT Action History} & \multicolumn{2}{c}{Preds Action History} \\
 & Top-1/5 Acc & Recall@5 & Top-1/5 Acc & Recall@5 \\\midrule
SA (Baseline) & 61.83 / 89.64 & 51.37 & 44.66 / 82.87 & 46.59 \\\midrule
CA (Q = Visual) & 37.94 / 81.79 & 40.00 & - & - \\
CA (Q = Text) & 64.03 / 94.24 & 64.70 & 40.42 / 78.71 & 49.48 \\
Residual CA (Sum) & 64.83 / 94.64 & 64.28 & 40.74 / 79.03 & 50.54 \\
Residual CA (Mean) & 64.39 / 94.44 & 63.81 & - & - \\
Residual CA (Product) & \underline{64.95 / 94.28} & 64.53 & 40.10 / 79.47 & 49.79 \\\midrule
Residual CA (Gated) & 63.55 / 94.20 & 63.50 & - & - \\
Cross-Residual & 57.90 / 89.80 & 58.94 & - & - \\
Gated Cross-Residual & \underline{61.14 / 92.40} & 56.02 & - & - \\\midrule
Bi-CA Concat (Independent) & 62.83 / 93.16 & 62.67 & - & - \\
Bi-CA Concat (Shared Weights) & 57.78 / 89.20 & 53.70 & - & - \\
Bi-CA (Sum) & 59.34 / 90.56 & 53.11 & - & - \\
Bi-CA (Mean) & 60.02 / 91.04 & 59.49 & - & - \\
Bi-CA (Product) & 62.18 / 91.28 & 55.50 & - & - \\
Bi-CA (Gated) & \underline{64.51 / 94.28} & 62.30 & \textbf{45.82 / 83.27} & \textbf{52.64} \\
Bi-CA (SA) & 61.14 / 90.36 & 56.53 & - & - \\
 \bottomrule
\end{tabular}
\end{minipage}
\hfill
\begin{minipage}{0.4\textwidth}
\centering
\setlength{\tabcolsep}{3pt}
\caption{Comparison of RGB and depth feature extractors on IKEA-ASM.}
\label{tab:feature_extractor}
\begin{tabular}{lllrr}
\toprule
\textbf{Feat. Extractor} & \textbf{Depth} & \textbf{Text} & \textbf{Top-1/5 Acc} & \textbf{Recall@5} \\\midrule
DINOv2 1B & - & - & 34.37 / 83.43 & 43.71 \\
DINOv3 0.8B & - & - & 31.29 / 82.35 & 39.25 \\
DINOv3 7B & - & - & \underline{37.37 / 87.64} & 46.57 \\\midrule
DINOv2 1B & DAv2 & - & 38.82 / 86.19 & 46.52 \\
DINOv3 7B & DAv2 & - & 38.18 / 86.95 & 48.75 \\
DINOv3 7B & DA3 & - & \underline{39.02 / 86.87} & 48.54 \\\midrule
DINOv2 1B & DAv2 & GT & 64.51 / 94.28 & 62.30 \\
DINOv3 7B & DAv2 & GT & \underline{65.87 / 94.20} & 64.05 \\
DINOv3 7B & DA3 & GT & 64.03 / 94.12 & 62.64 \\\midrule
DINOv2 1B & DAv2 & Preds & 45.82 / 83.27 & 52.64 \\
DINOv3 7B & DAv2 & Preds & \textbf{48.90 / 85.11} & 53.01 \\
DINOv3 7B & DA3 & Preds & 48.58 / 84.55 & 53.22 \\
\bottomrule
\end{tabular}
\end{minipage}

\end{table*}

\vspace{0.1cm}
\noindent \textbf{Cross-Residual Attention.}
To further explore lightweight cross-modal interactions, we consider \emph{cross-residual attention} modules, which replace the attention operator with simple feature mixing between modalities. Given the attended output $X'$ obtained from a directional cross-attention step, cross-residual fusion is defined as:
\begin{equation}
Y = X_\text{tgt} + X'.
\end{equation}
This additive shortcut enables a direct injection of complementary cues without requiring a full attention layer.

In practice, we evaluate only the most informative configuration from the previous analysis, namely $\text{CA} (Q = X_t)$. In this case, the textual queries attend to the visual features, producing an attended representation, $X_t'$, that is then injected into the visual stream via
\begin{equation}
Y = X_v + X_t'.
\end{equation}
This applies the residual fusion in the direction that gave the strongest cross-modal signal under ground-truth action histories.

Although the gated variant yields a modest improvement over the standard cross-residual fusion, both approaches perform worse than the self-attention baseline under ground-truth action histories. Consequently, neither configuration was selected for evaluation under predicted action histories.

\vspace{0.1cm}
\noindent \textbf{Bidirectional Cross-Attention.}
While the previous modules rely on a single direction of interaction, bidirectional cross-attention (Bi-CA) allows both modalities to attend to one another. This produces a pair of cross-attended representations
\begin{equation}
(X_v',\,X_t') = \bigl(\text{CA}(X_v,X_t,X_t),\ \text{CA}(X_t,X_v,X_v)\bigr),
\end{equation}
where $\text{CA}(Q,K,V)$ denotes cross-attention with queries $Q$, keys $K$, and values $V$. $X_v'$ enriches the visual stream with textual cues, and $X_t'$ injects visual information into the textual embedding.

We evaluate two projection strategies. In the \emph{independent} variant, each direction uses its own set of projection matrices, allowing the model to learn asymmetric transformations for text-to-visual and visual-to-text attention. In contrast, the \emph{shared} variant ties the projection weights across directions, enforcing a symmetric interaction between modalities. Although computationally lighter, the shared formulation consistently underperforms its independent counterpart, suggesting that visual and text require different attention parametrizations.

Several fusion operators are applied to combine the two attended representations, including concatenation, element-wise sum, mean fusion, embeddings product, and self-attention. Among all variants, the gated fusion achieves the strongest performance. This operator computes
\begin{equation}
Y = g \odot X_v' + (1 - g) \odot X_t',
\end{equation}
where $g \in [0,1]$ is a learned gating vector that adaptively balances the contribution of each modality. This flexibility enables the model to downweight unreliable textual inputs under predicted action histories, leading to the best overall performance among all evaluated fusion strategies, even under the realistic setting.

\subsection{Feature Extraction}
\noindent In terms of feature extraction, we continue to use DistilBERT~\cite{sanh2019distilbert} as the text encoder due to its strong unimodal performance reported in~\cite{aag}. Errors in the textual stream largely stem from the inherent ambiguity of anticipating the next action based on a short action history, rather than from limitations of the encoder itself. For this reason, our analysis focuses primarily on the RGB and depth modalities.

To assess the impact of visual feature quality, we extend the evaluation to include DINOv3~\cite{simeoni2025dinov3} (7B and 0.8B variants) and Depth Anything 3 (DA3)~\cite{lin2025depth}, comparing them against the original DINOv2~\cite{oquab2023dinov2} and Depth Anything v2 (DAv2)~\cite{NEURIPS2024_26cfdcd8}. Table~\ref{tab:feature_extractor} reports results across the following settings: RGB-only, RGB-Depth, RGB-Depth-Text under ground-truth action histories, and RGB-Depth-Text under predicted action histories. The latter represents the realistic operating regime of our method and constitutes the primary metric of interest.

\begin{figure}[t]
    \centering
    \includegraphics[width=0.9\linewidth]{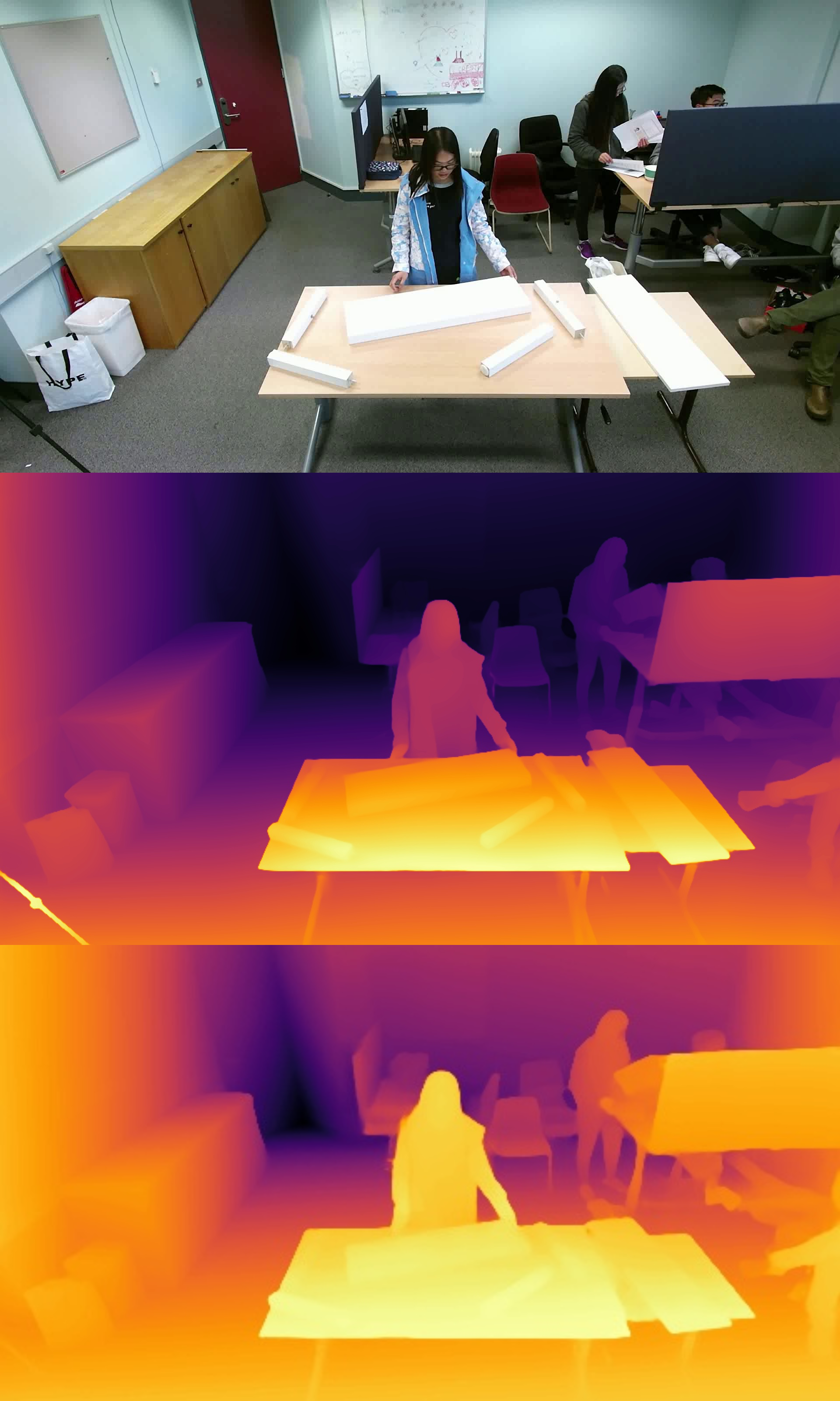}
    \caption{Qualitative comparison of depth frames generated by DAv2 (middle) and DA3 (bottom), with the RGB input shown on top for reference. Note how bjects on the table are more clearly distinguished with DAv2 than with DA3.}   
    \label{fig:depth_comp}
\end{figure}

In the RGB-only setting, DINOv3~7B consistently outperforms both the smaller 0.8B variant and DINOv2 across Top-1, Top-5, and Recall@5 metrics. This trend holds despite evaluating the 7B model in FP16 due to computational constraints, which slightly reduces numerical precision.

When introducing depth, we observe a slight decrease in performance when using DINOv3 with DAv2, whereas DA3 provides performance slightly above the DINOv2-DAv2 baseline. Despite this, when depth is combined with textual features, DAv2 yields the strongest performance under both ground-truth and predicted action histories. We attribute this to DAv2 preserving finer geometric details, as illustrated qualitatively in Figure~\ref{fig:depth_comp}. Given its superior performance in the multimodal configuration, we adopt DINOv3~7B with Depth Anything v2 in all subsequent experiments.

\subsection{Blurriness Analysis}
\noindent The blurriness-based keyframe selection strategy relies on a dataset-specific Laplacian variance threshold to identify visually reliable frames within a short temporal window of five frames. Since absolute Laplacian variance values depend strongly on camera motion, viewpoint, and scene structure, a single global threshold is insufficient across datasets.

Laplacian variance serves as an effective proxy for frame sharpness, with higher values indicating stronger edge responses and reduced motion blur. Empirically, values above 300 correspond to very sharp frames for which the most recent observation is already visually reliable. The range $100-300$ represents sharp and informative frames, where multiple candidates within the temporal window often satisfy the quality criterion. Values between 50 and 100 indicate moderate blur, motivating evaluation of nearby frames, while values below 50 correspond to pronounced motion blur. Frames with Laplacian variance below 30 are severely blurred, offering limited usable structure and forcing the strategy to default to the most recent frame when no sharper alternative exists.

Table~\ref{tab:laplacian} reports the empirical distribution of Laplacian variance values and the resulting frame selection behavior, including the percentage of updated frames and cases where all frames within the temporal window fall below the threshold (\textit{Too Blurry)}. A threshold of 100 was chosen as default, providing robust sharpness across datasets. IKEA-ASM exhibits consistently high sharpness (mean 301.2, minimum 128.2), such that the threshold never triggers frame replacement. Meccano, in contrast, features egocentric motion and frequent blur (means 25.7-34.1), requiring additional evaluation at thresholds of 50 and 30. A threshold of 50 balances frame updates while preserving reasonable quality. Assembly101 lies in an intermediate regime (mean 101.4), where the strategy reliably identifies sharper frames within the five-frame window, yielding a large number of modified selections (20.6\%) while minimizing cases where all frames are too blurry (11.7\%). Frames outside these percentage ranges correspond to instances where the last frame prior to anticipation already exceeds the threshold and therefore requires no modification.

\begin{figure}
    \centering
    \includegraphics[width=\linewidth]{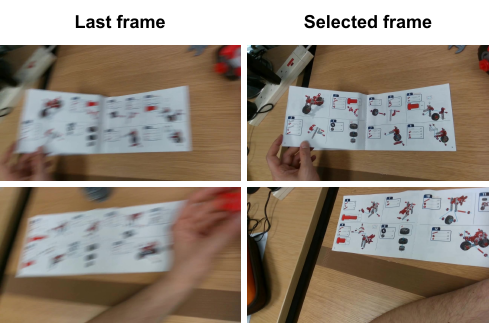}
    \caption{Sample frames from the Meccano dataset illustrating motion blur caused by the egocentric head-mounted camera (left) and the corresponding frames selected by our keyframe selection method (right).}
    \label{fig:blur_qualitative}
\end{figure}

This analysis is further supported by qualitative results on the Meccano dataset, which exhibits the lowest sharpness values due to its egocentric camera setup. Figure~\ref{fig:blur_qualitative} presents qualitative examples of both selected and original frames, further demonstrating the effectiveness of the proposed method within the AAG+ framework in identifying and selecting the most informative frames.

\begin{table}[t]
\centering
\caption{Laplacian variance statistics and frame selection behavior across datasets. }
\label{tab:laplacian}
\resizebox{\linewidth}{!}{%    
\begin{tabular}{lrrrrrr}
\toprule
\textbf{Dataset} & \textbf{Threshold} & \textbf{Mean} & \textbf{Min} & \textbf{Max} & \textbf{Updated (\%)} & \textbf{Too Blurry (\%)} \\
\midrule
IKEA-ASM      & 100 & $301.2 \pm 125.3$ & 128.2 & 801.4 & 0.0     & 0.0 \\\midrule
\multirow{3}{*}{Meccano}       & 100 & $34.1 \pm 18.6$ & 1.9  & 169.8 & 1.3  & 97.2 \\
       & 50  & $29.2 \pm 17.2$ & 1.9  & 169.8 & 13.1 & 67.9 \\
       & 30  & $25.7 \pm 15.6$ & 1.9  & 167.5 & 24.9 & 20.6 \\\midrule
Assembly101   & 100 & $101.4 \pm 15.1$ & 71.8 & 212.1 & 20.6 & 11.7 \\
\bottomrule
\end{tabular}
}
\end{table}

\section{Performance report of the action recognizer to generate action 
history}\label{app:ar_results}

\noindent As described in the main paper, AAG+ constructs the action history using an action recognizer with the same architecture, adapted to each recognition benchmark. All experiments are conducted using the same modalities employed for action anticipation. Baselines for comparison are dataset-specific: IKEA-ASM uses I3D~\cite{carreira2017quo} (Top-5 not provided), Meccano uses SlowFast~\cite{Feichtenhofer_2019_ICCV}, and Assembly101 uses TSM~\cite{lin2019tsm}, trained on all fixed cameras. Action recognition results are reported in Table~\ref{tab:single-frame-ar}.

The proposed method outperforms the previous single-frame approach, AAG, on both IKEA-ASM and Meccano, and even surpasses some video-based methods on IKEA-ASM. On Assembly101, the model benefits from improved fusion of RGB and depth features, achieving higher performance when depth is included. However, results remain slightly below those of the original AAG and notably below video-based approaches. We attribute this gap to the increased importance of accurately modeling action history in more complex scenarios, as evidenced by the multimodal analysis of AAG+ in the anticipation setting, where longer action sequences and visually ambiguous intermediate states reduce the reliability of single-frame visual cues.

Importantly, despite lower single-frame action recognition accuracy compared to video-based methods, AAG+ achieves stronger action anticipation performance, as reported in the main paper. This improvement stems from the proposed blurriness-based keyframe selection and robustness mechanisms for handling imperfect action histories, enabling AAG+ to outperform both AAG and several video-based approaches in anticipation.

\begin{table*}
    \begin{minipage}{0.55\textwidth}
    \centering
    \caption{Single-frame action recognition performance (Top-1/5).}
    \label{tab:single-frame-ar}
    \begin{tabular}{lccc}
    \toprule
    \textbf{Methods} & \textbf{IKEA-ASM} & \textbf{Meccano} & \textbf{Assembly101} \\ \midrule
    Video (RGB-only) & 57.58 / - & 45.16 / 73.75 & - \\
    Video (Multimodal) & 64.25 / - & \textbf{49.66 / 73.75} & \textbf{43.60} / - \\
    AAG$_{AR}$ (RGB, AH) & 66.43 / 92.56 & 31.21 / 66.03 & 34.19 / 58.30 \\
    AAG$_{AR}$ (RGB-D, AH) & 66.84 / 92.29 & 31.14 / 66.52 & 32.77 / 56.19 \\
    AAG+$_{AR}$ (RGB, AH) & 72.38 / 94.58 & 33.19 / 71.34 & 30.74 / 57.97  \\ 
    AAG+$_{AR}$ (RGB-D, AH) & \textbf{72.95 / 95.04} & 33.79 / 73.29 & 31.49 / 59.09 \\ 
    \bottomrule
    \end{tabular}
    \end{minipage}
    \hfill
    \begin{minipage}{0.45\textwidth}
        \centering
    \caption{Computational comparison.}
    \label{tab:compute_comparison}
    \begin{tabular}{lcc}
        \toprule
        \textbf{Method} & \textbf{\begin{tabular}[c]{@{}c@{}}Trainable\\  Params (M)\end{tabular}} & \textbf{\begin{tabular}[c]{@{}c@{}}Processed\\ Frames\end{tabular}} \\
        \midrule
        AVT~\cite{AVT} & 392 & 10 \\
        TempAgg~\cite{TempAgg} & 123 & 37 \\
        RULSTM~\cite{RULSTM} & 67 & 14 \\
        VLMAH~\cite{Manousaki_2023_ICCV} & 45 & 8 \\
        AAG~\cite{aag} & \textbf{24} & \textbf{1} \\
        \textbf{AAG+} & 34 & \textbf{1} \\
        \bottomrule
    \end{tabular}
    \end{minipage}
\end{table*}

\section{Computational Analysis}\label{app:comp_analysis}

\noindent To further highlight the computational advantages of AAG+ over existing approaches, we report a comparative analysis in Table~\ref{tab:compute_comparison}. For each state-of-the-art method considered in the main paper, we list the number of trainable parameters and the number of processed input frames. Feature extraction parameters are excluded, as feature encoders are frozen across all methods evaluated on pre-extracted features. Thus, the analysis focuses on the core model components and their input requirements.

AAG+, like its single-frame predecessor AAG, achieves competitive performance while incurring substantially lower computational overhead. Methods such as AVT and TempAgg require significantly more trainable parameters, whereas RULSTM is more compact but delivers lower performance. VLMAH is closer to AAG+ in model size, however, it relies on fine-tuned video feature extractors, which restricts training flexibility and limits adaptability across domains. In contrast, AAG+ leverages self-supervised features without task-specific fine-tuning.

Furthermore, video-based approaches process between 8 and 37 frames per sample, resulting in a proportional increase in feature extraction cost per video clip compared to AAG+’s single-frame design. While the enhanced mechanisms in AAG+ slightly increase the number of trainable parameters relative to AAG, this modest overhead yields substantial performance gains over video-based methods and enables more effective use of the available modalities.

\section{Extended Multimodal Analysis}\label{sec:multi-analysis}
\noindent In this appendix, we extend the modality contribution analysis presented in Section~V to the other two modality permutations: RGB-Depth and RGB-Text, focusing on the complementary roles of RGB appearance with these contextual cues. While the main paper analyzes the interaction between visual and textual modalities, here we examine (a) the specific interaction between the single-frame RGB input and the procedural memory (textual modality), and (b) how different visual sources contribute to correct and incorrect anticipation decisions at a per-sample level.

\vspace{0.1cm}
\noindent \textbf{RGB-Text Analysis.} While Section~V.B analyzes interactions between visual (RGB-Depth) and textual modalities, here we focus on RGB when examining visual-textual complementarity. RGB serves as the primary source of semantic and object-level information, whereas depth acts mainly as a supportive geometric cue that is most effective when (a) is fused with RGB and rarely functions as an independent predictor, and (b) under static, exocentric camera setups (see subsequent analysis). Isolating RGB therefore enables a clearer analysis of how instantaneous visual semantics interact with long-term procedural context.

\begin{figure*}[t]
    \centering
    \includegraphics[width=\linewidth]{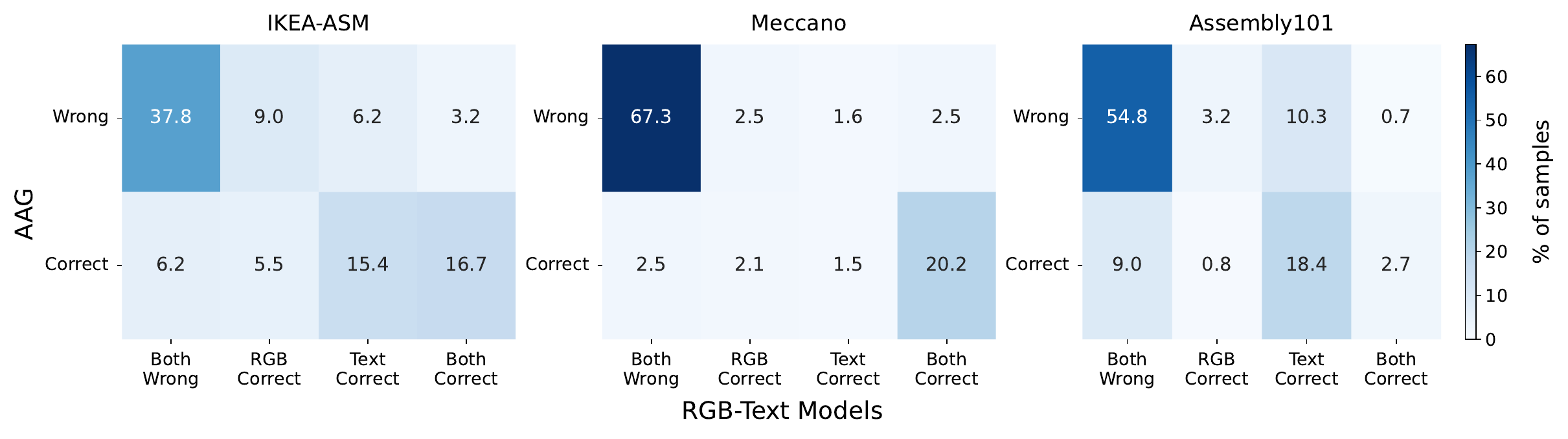}
    \includegraphics[width=\linewidth]{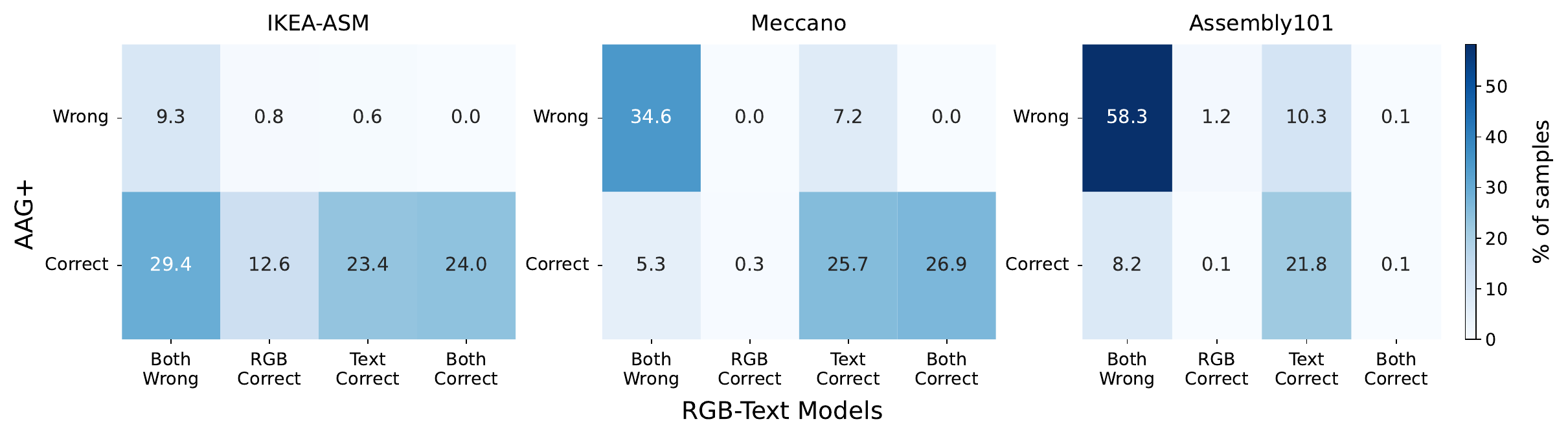}
    \caption{Confusion matrix-style modality breakdown for AAG and AAG+ across datasets under RGB-Text fusion.}
    \label{fig:analysis_rgb_text}
\end{figure*}

The RGB-Text meta-analysis (Figure~\ref{fig:analysis_rgb_text}) reveals a clear dataset-dependent shift in modality reliance in AAG+. On IKEA-ASM, which features shorter procedural sequences and stable third-person viewpoints, AAG+ achieves a balanced contribution between visual and textual cues, with a substantial increase in jointly correct predictions. This indicates effective cross-modal integration, where single-frame appearance is reliably grounded by action-history semantics.

In contrast, Meccano exhibits a strong dominance of the textual modality: the vast majority of correct predictions arise from text-only or joint RGB-Text correctness, while RGB-only correctness is negligible. This reflects the fine-grained, visually repetitive nature of the task, where instantaneous visual evidence is often insufficient to disambiguate the next action without procedural context. The trend is most evident on Assembly101, where correct predictions are almost exclusively driven by the textual action-history encoder. The near absence of RGB-only and joint correctness confirms that, in long-horizon, highly variable procedural tasks, single-frame visual cues provide limited predictive power. Compared to AAG, AAG+ further consolidates this behavior, indicating that its refinements enable more decisive reliance on semantic temporal priors rather than noisy visual evidence.

\vspace{0.1cm}

\begin{figure*}[t]
    \centering
    \includegraphics[width=\linewidth]{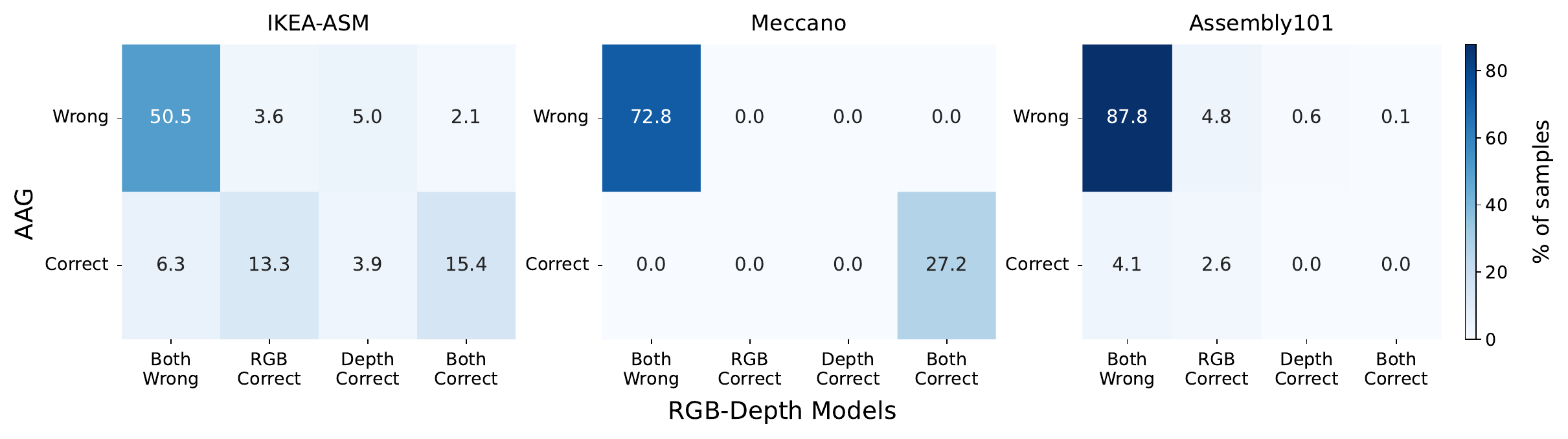}
    \includegraphics[width=\linewidth]{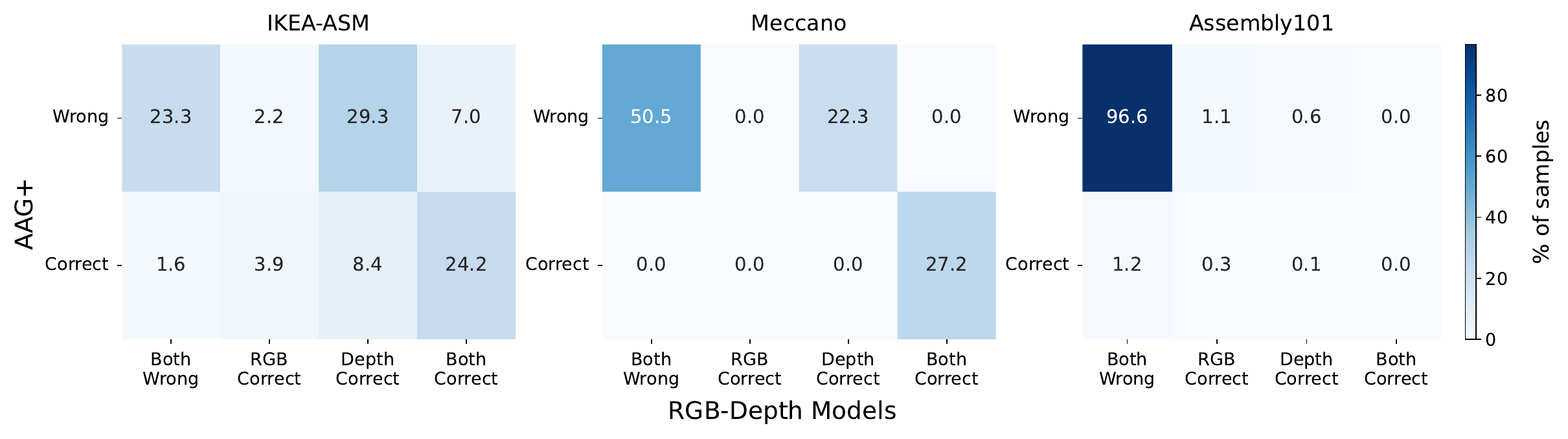}
    \caption{Confusion matrix-style modality breakdown for AAG and AAG+ across datasets under RGB-Depth fusion.}
    \label{fig:RGB-Depth}
\end{figure*}

\noindent \textbf{Visual Analysis.} An RGB-Depth meta-analysis reveals clear differences in how visual cues are exploited in AAG and AAG+. The visual source contribution confusion-style analysis, shown in Figure~\ref{fig:RGB-Depth} reveals that the contribution of geometric cues is strongly dataset-dependent and closely tied to capture conditions and task structure.

On IKEA-ASM, which is recorded from a static third-person viewpoint with limited camera motion, depth estimates are comparatively reliable and stable. Under these conditions, AAG+ more effectively exploits geometric information than AAG, as evidenced by an increased proportion of samples in which both RGB and Depth lead to correct predictions. This shift indicates improved cross-modal alignment, where depth reinforces appearance-based cues rather than acting as a noisy auxiliary signal. In contrast, for both Meccano and Assembly101, the analysis indicates that the nature of the datasets themselves necessitates a shift away from visual dominance. In Meccano, visual modalities appear largely redundant: when the visual branch yields a correct prediction, both RGB and Depth predictors typically arrive at the decision independently. Similarly, in Assembly101, AAG+ exhibits a reduction in isolated RGB- or Depth-correct cases, relegating visual cues to a secondary role. Consequently, this verifies for both datasets, any significant performance increase in AAG+ is attributed to, and necessitated by, enhanced reliance on the textual modality, as the visual ambiguity and procedural complexity inherent to these domains limit the ceiling of pure visual processing.

This underscores a fundamental characteristic of procedural activity analysis: as the action space grows and action sequences become more variable, memory of past actions becomes increasingly critical for accurate anticipation. Due to the inherent visual complexity and occlusion in these scenes, specific visual information extracted from a single frame is frequently ambiguous. Thus, the model must rely on the temporal scaffold provided by the action history to disambiguate the future state.

\section{Qualitative Comparison of VLMs for Action History Retrieval}\label{app:vlms}
\begin{figure*}[t]
    \includegraphics[width=\textwidth]{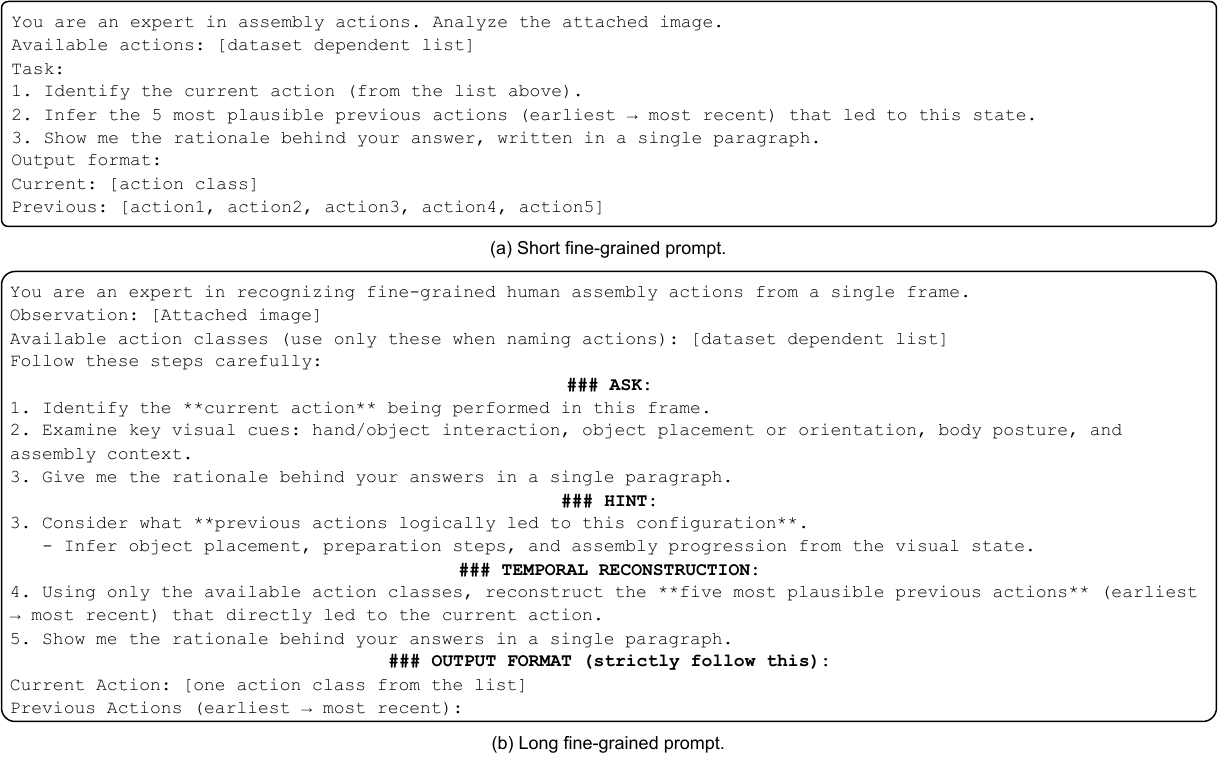}
    \caption{Fine-grained prompts for action history retrieval. (a) Short version, which concisely summarizes past actions for efficient retrieval. (b) Long version, which includes richer contextual details to support more accurate retrieval. When action sequences are available, they are provided as an attached file, and the model is explicitly informed of its existence.}
    \label{fig:prompts}
\end{figure*}

\noindent We conduct a qualitative evaluation of VLMs to gain deeper insights into their behavior across different prompt formulations, task complexities, and viewpoints. Figure~\ref{fig:prompts} illustrates the two prompt variants used in the main paper, short and long fine-grained prompts, designed to assess the impact of instruction granularity. We then report results for each dataset on frames where the single-frame action recognizer performs well. For each example, we present: (1) the current action, (2) the previous five actions, and (3) the model’s rationale for predicting these actions. To facilitate analysis, correct observations are highlighted in green, incorrect ones in red, and model assumptions are shown in yellow.

We should note that we additionally experimented with Llama-4 Maverick under the same prompt formulations. However, its outputs were consistently less coherent and substantially less aligned with the ground-truth action histories than both GPT 5.1 and Gemini 3 Flash. For this reason, we omit qualitative examples for Llama-4 Maverick.

\begin{figure*}[p]
    \includegraphics[width=\textwidth]{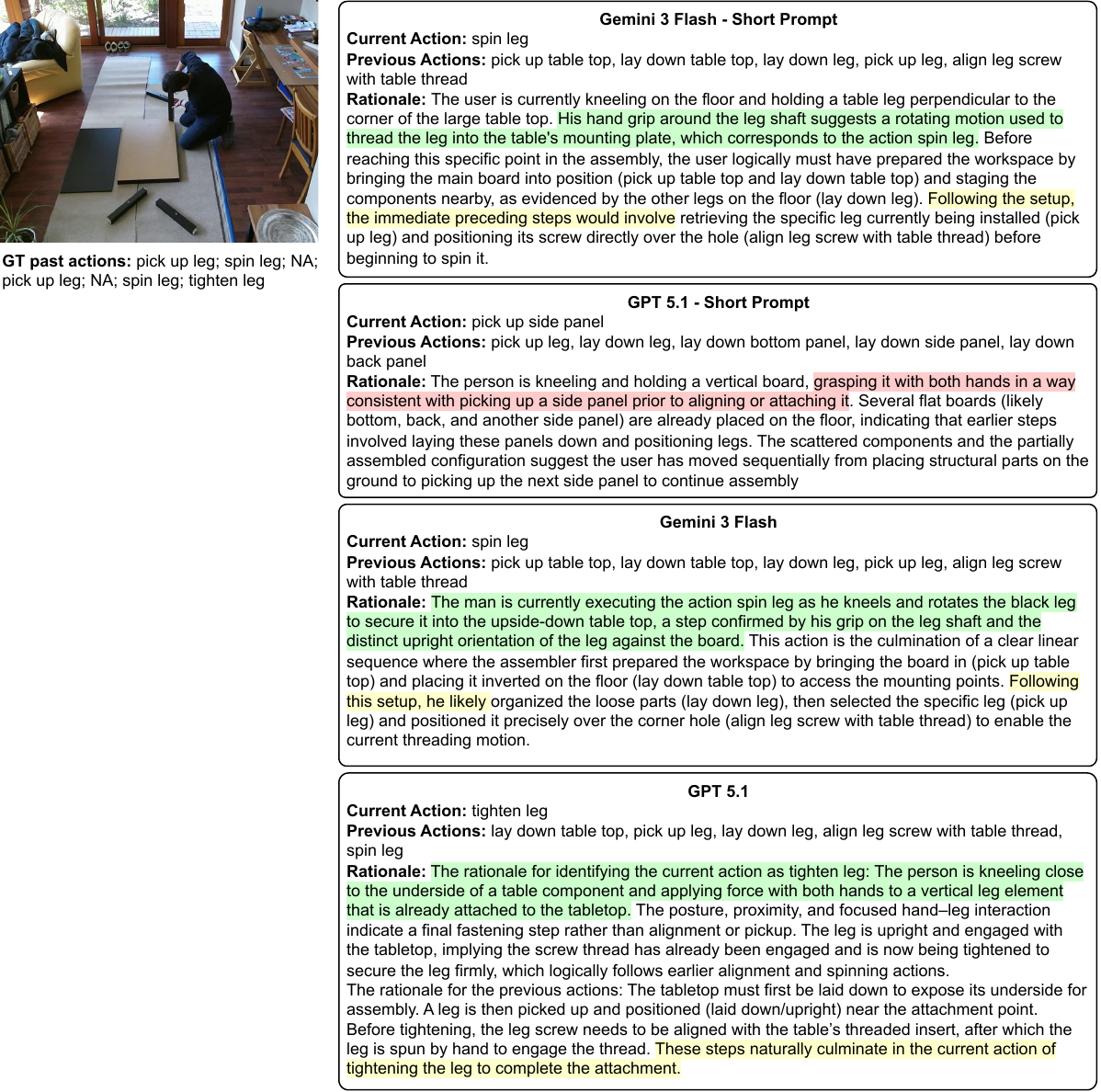}
    \caption{VLMs qualitative results on IKEA-ASM dataset. Note the colors noting correct observations (green), incorrect (red) or assumed (yellow).}
    \label{fig:ikea_vlms}
\end{figure*}

The behavior observed on the IKEA-ASM dataset, shown in Figure~\ref{fig:ikea_vlms}, is consistent with the quantitative evaluation. Gemini 3 Flash generally outperforms GPT-5.1, which, under fine-grained prompts, often fails to correctly identify the ongoing action and consequently mispredicts past actions. We further observe that the VLMs sometimes rely on prior knowledge or assumptions rather than visual evidence of the current state and objects. VLMs still lack precise moment-level information, highlighting both the effectiveness of the action recognizer for this task and the limitations of current language-based models for this task.

\begin{figure*}[p]
    \includegraphics[width=\textwidth]{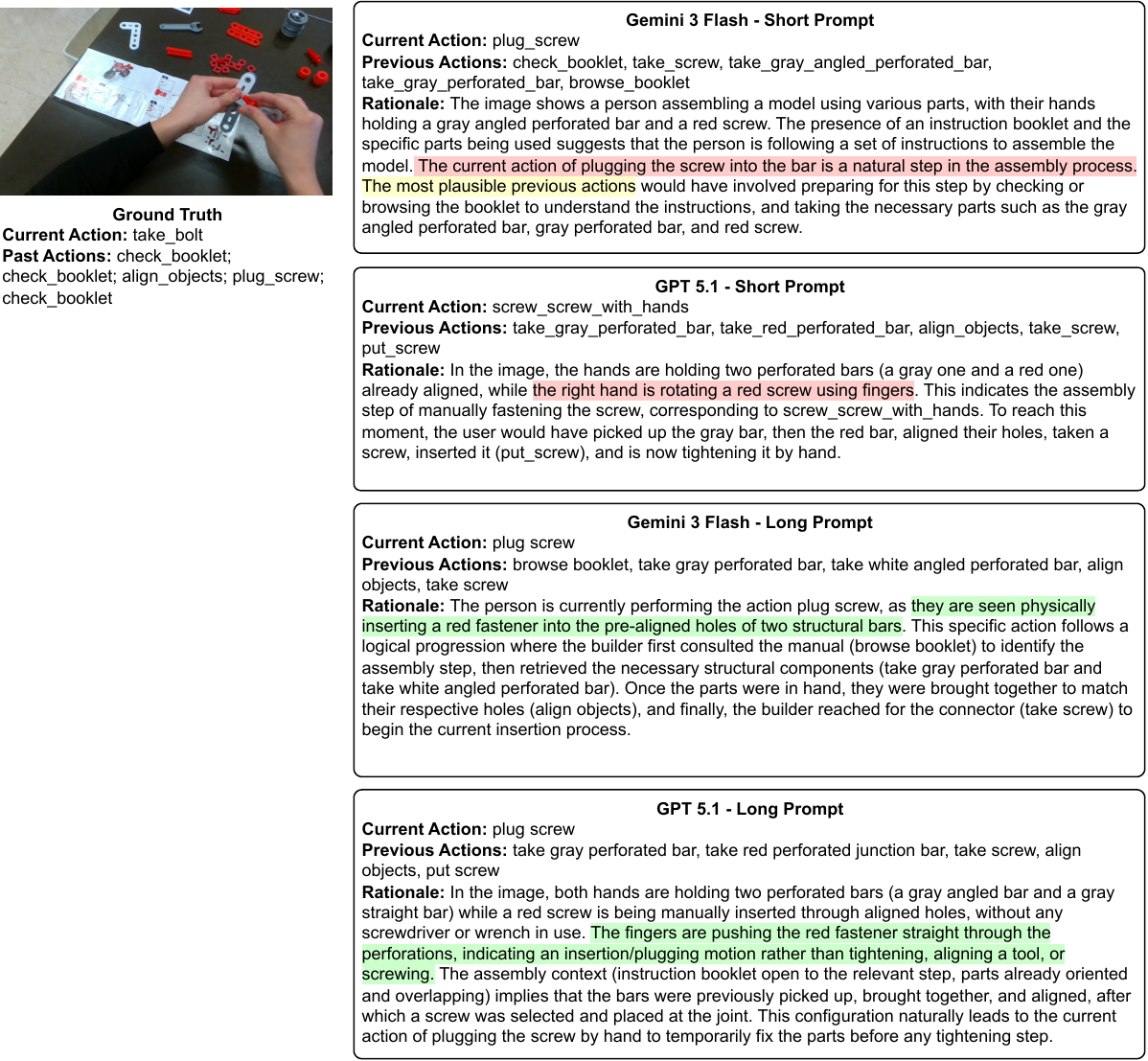}
    \caption{VLMs qualitative results on Meccano dataset. Note the colors noting correct observations (green), incorrect (red) or assumed (yellow).}
    \label{fig:meccano_vlms}
\end{figure*}

Figure~\ref{fig:meccano_vlms} presents the qualitative results on the Meccano dataset. In contrast to the previous dataset, partial observation of an action in this setting negatively affects VLM behavior: when only the final frames of the \textit{take\_bolt} action are visible, the visual evidence leads the vision-language models to predict \textit{plug\_screw} instead. This example further highlights the limitations of these models, which, despite being significantly heavier than the action recognizer, fail to match its performance. Notably, the explicit action history leveraged by the action recognizer, or the temporal modeling employed by video-based methods, helps disambiguate such cases and enforces the logical ordering of actions: a bolt cannot be plugged unless it has first been taken.

\begin{figure*}[p]
    \includegraphics[width=\textwidth]{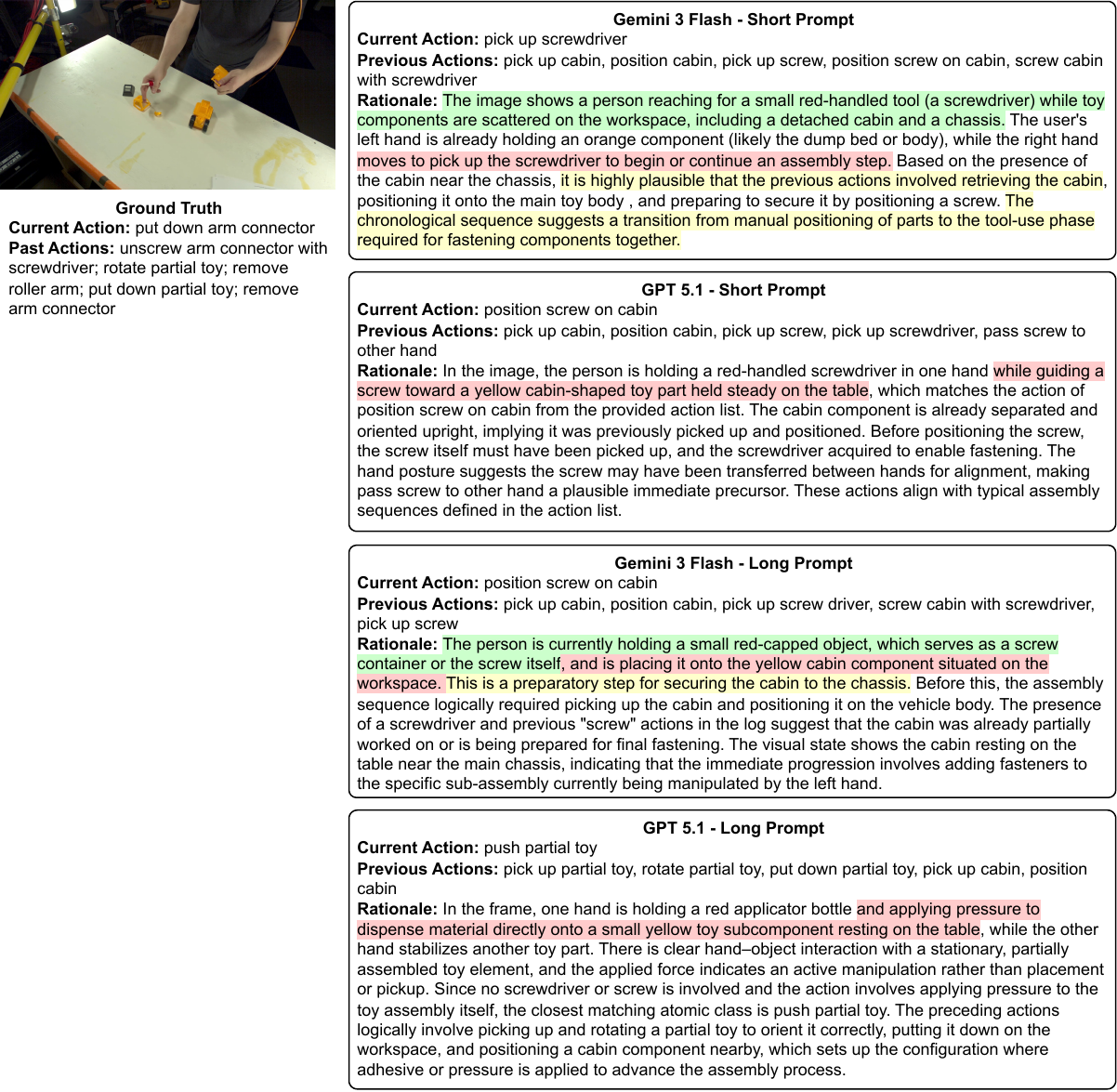}
    \caption{VLMs qualitative results on Assembly101 dataset. Note the colors noting correct observations (green), incorrect (red) or assumed (yellow).}
    \label{fig:assembly_vlms}
\end{figure*}

Finally, qualitative results on the Assembly101 dataset are presented in Figure~\ref{fig:assembly_vlms}. This dataset includes both assembly and disassembly processes, which pose a significant challenge for VLMs. Notably, Gemini produces more coherent results than GPT, which often relies on erroneous visual observations. However, even when Gemini correctly interprets the current observation, assumptions about an assembly (rather than a disassembly) process  lead to completely incorrect predictions for previous actions.

%\end{comment}

\end{document}